\documentclass[11pt]{article}
\usepackage{listings}
\usepackage{multirow}
\usepackage[most]{tcolorbox}
\usepackage{setspace}
\usepackage{siunitx}
\tcbset{
  promptbox/.style={
    enhanced,
    colback=white,
    colframe=black,
    boxrule=0.4pt,
    arc=2pt,
    left=6pt,
    right=6pt,
    top=6pt,
    bottom=6pt,
    boxsep=0pt,
    breakable,
    fontupper=\ttfamily\small,
  }
}
\usepackage{adjustbox}
\usepackage{graphicx}
\usepackage{authblk}
\usepackage[]{acl}
\usepackage{booktabs}
\usepackage{times}
\usepackage{latexsym}
\usepackage{graphicx}
\usepackage{float}      
\usepackage{dblfloatfix} 
\usepackage[T1]{fontenc}
\usepackage{arabtex}

\usepackage[utf8]{inputenc}
\usepackage{enumitem}
\usepackage{microtype}

\usepackage{inconsolata}

\usepackage{graphicx}

\def\XXX#1{{\textcolor{red}{XXX #1}}}
\setcode{utf8}
\def\XXX#1{}

\title{Current state of LLMs for Arabic dialectal machine translation}

\author{
  Josef Jon$^{1}$ \quad Rawan Bondok$^{2}$ \quad Ondřej Bojar$^{1}$ \\
  $^{1}$Charles University, Faculty of Mathematics and Physics, Prague, Czechia \\
  \texttt{jon@ufal.mff.cuni.cz} \\
  $^{2}$TU Darmstadt, Germany \\
  \texttt{rawanessam34@gmail.com}
}

\begin{document}
\maketitle\begin{abstract}
This work presents an evaluation of large language models (LLMs) for English to dialectal Arabic machine translation on the MADAR dataset. We evaluate both translation directions (English to Arabic and vice-versa) on 16 Arabic dialects. Our experiments cover a diverse set of models, including specialized Arabic models (Jais, Nile), multilingual models (Gemma, Command-R, Mistral, Aya), and commercial APIs (GPT-4.1). We employ multiple evaluation metrics: BLEU, CHRF, COMET (both reference-based and reference-less variants) and GEMBA (LLM-as-a-judge), as well as a small-scale manual evaluation, to assess translation quality. We discuss the challenges of automatic MT evaluation, especially in the context of Arabic dialects. We also evaluate the ability of LLMs to classify the dialect used in a text. The study offers insights into the capabilities and limitations of current LLMs for dialectal Arabic machine translation, particularly highlighting the difficulty of handling dialectal diversity, although the results may be influenced by possible training data contamination, which is always a concern with LLMs.
\end{list} 
\end{abstract}

\section{Introduction}

Machine translation (MT) for dialects of Arabic remains a challenging problem despite significant progress in neural and large language model (LLM)–based approaches. Arabic is characterized by a high degree of linguistic complexity, including rich morphology, flexible word order, and diglossia between Modern Standard Arabic (MSA) and numerous regional dialects. While MSA is widely used in formal writing and media, everyday communication across the Arab world primarily relies on dialectal varieties, which differ substantially both from MSA and from each other. 
Moreover, even in a dialect spoken in a single country, there can be large regional variations. 

In the last few years, we have witnessed rapid advances in multilingual LLMs, which have demonstrated strong performance on a wide range of natural language processing tasks, including MT \cite{kocmi-etal-2025-findings}. However, their capabilities for dialectal Arabic are still insufficiently understood. Most existing evaluations \cite{saeed2025machine} focus on MSA, leaving open questions about how well current LLMs handle the full spectrum of Arabic dialectal variation.

Dialectal Arabic MT poses unique challenges beyond data scarcity. Dialects show substantial lexical, phonological, and syntactic variation and a lack of standardized orthography. 
Even within a single country, there can be considerable variations depending on region, speaker, and domain. These factors complicate both model training and evaluation. In particular, automatic evaluation metrics that rely on a single reference translation may fail to capture valid alternative renderings, leading to unreliable quality estimates. While this is a general problem of reference-based metrics, it is especially pronounced for dialectal Arabic, where multiple translations may be equally acceptable but differ significantly at the surface form.  \XXX{@Rawan: Do you think the last sentence is true and that it is worded clearly? That reference-based metric might be more suitable for e.g. English than Arabic dialects, because English is more standardized/offers less ways to formulate the same though? I agree yes this true I will add an example from data to support this}

We present a comprehensive evaluation of 16 large language models for Arabic$\leftrightarrow$English machine translation across 16 Arabic dialects using the MADAR test set~\cite{bouamor-etal-2018-madar}.  We evaluate both translation directions and cover a diverse set of models, including Arabic-specialized LLMs, strong multilingual models, and commercial API-based systems. We employ BLEU, ChrF, COMET and LLM-as-a-judge metrics, as well as a small-scale manual evaluation. Through quantitative and qualitative analysis, we investigate model performance patterns across dialects, translation directions, and evaluation metrics, shedding light on the current capabilities and limitations of LLMs for Arabic MT. 
Our findings show both the progress made by recent LLMs and the persistent challenges posed by dialectal diversity. By providing a systematic comparison, this study aims to serve as a basis for selecting LLMs for MT in various scenarios, as a reference point for future research on Arabic machine translation and as an encouragement for development of more robust models and evaluation methodologies tailored to dialectal Arabic.
We note that due to the unavailability of the training data for many of the models, we do not know if the MADAR test set was used during the training of the  LLMs. This is a major shortcoming of this work, which must be taken into account when basing any decisions on our results.


The main contributions of this work are:
\begin{itemize}
    \item A comprehensive evaluation of 16 LLMs on Arabic-English translation across 16 dialects
    \item Manual analysis of selected examples
    \item Analysis of automatic evaluation metrics for dialectal Arabic MT
    \item Insights into dialect-specific challenges and model performance patterns
    \item Public release of results and analysis scripts\footnote{\url{https://github.com/cepin19/arabic_llms}}
\end{itemize}

The remainder of this paper is organized as follows: Section~\ref{sec:related} reviews related work. Section~\ref{sec:data} describes the MADAR dataset and  the evaluated models. Section~\ref{sec:experiments} presents experimental settings and evaluation metrics. Section~\ref{sec:results} presents and analyzes results. Section~\ref{sec:discussion} discusses  the implications of the results. Section~\ref{sec:conclusion} summarizes our findings.

\section{Related Work}
\label{sec:related}

The work on Arabic MT copies the general trends in the field of MT. Early work relied on rule-based and phrase-based statistical MT systems, often combined with morphological analysis \cite{habash2009arabic,habash-hu-2009-improving}. 

In recent years, neural machine translation (NMT), particularly Transformer-based architectures \cite{vaswani}, has become the dominant paradigm for all MT, including Arabic \cite{almahairi2016resultarabicneuralmachine,durrani-etal-2016-qcris}. The newest iteration of neural network-based approaches, LLMs, have also been finetuned with success for Arabic translation \cite{nagoudi-etal-2022-arat5,nagoudi-etal-2022-turjuman}. A number of recent LLMs with Arabic capabilities are listed in Table \ref{tab:models}.

Other works have addressed the specific issue of dialectal Arabic MT, either through pivoting via MSA \cite{sajjad-etal-2013-translating,salloum-habash-2013-dialectal}, or by finetuning the models \cite{zbib-etal-2012-machine,unsupervised,nagoudi-etal-2021-investigating,Baniata2018ANM,HEAKL2024113}. 
The translation of Arabic dialects remains a low-resource problem due to limited parallel data and high linguistic variation. To address this, several datasets have been introduced, including MADAR, which covers dialects from multiple countries aligned with MSA and English \cite{bouamor-etal-2018-madar}.

One of the regular evaluation campaigns that covers Arabic dialects (Tunisian and Levantine) is IWSLT \cite{ahmad-etal-2024-findings,agostinelli-etal-2025-findings}. LLMs have proven to be effective in this setting as well \cite{ben-kheder-etal-2024-aladan}.

\section{Data and Models}
\label{sec:data}

\begin{table}[t]
\centering
\small
\begin{tabular}{l@{\hspace{5pt}}l}
\hline
\textbf{Model} & \textbf{Reference} \\
\hline
\multicolumn{2}{l}{\textit{Arabic-specialized models}} \\
Jais-2-8B-Chat &  \citet{jais2_2025} \\
Jais-2-70B-Chat & \citet{jais2_2025} \\
Nile-Chat-12B & \citet{shang-etal-2025-nile} \\
c4ai-command-r7b-arabic & \citet{alnumay2025command} \\
\hline
\multicolumn{2}{l}{\textit{Multilingual models}} \\
Aya Expanse 8B &  \citet{dang2024ayaexpansecombiningresearch} \\
Aya Expanse 32B & \citet{dang2024ayaexpansecombiningresearch} \\
c4ai-command-r-08-2024  & \citet{cohere_for_ai_2024} \\
c4ai-command-r-v01  & \citet{cohere_for_ai_r} \\
Command-A-Translate-08-2025 & \citet{kocmi-etal-2025-command} \\
Gemma-3-4B-IT & \citet{gemma_2025} \\
Gemma-3-27B-IT &  \citet{gemma_2025} \\
EuroLLM-9B-Instruct & \citet{martins2025eurollm9btechnicalreport} \\
Mistral-Small-3.2-24B-Instruct & \citet{Mistral-Team_2025} \\
Qwen3-4B-Instruct-2507 & \citet{yang2025qwen3technicalreport} \\
Llama-3.3-70B-Instruct &  \citet{grattafiori2024llama3herdmodels} \\
\hline
\multicolumn{2}{l}{\textit{Commercial API models}} \\
GPT-4.1-nano & \citet{openai2024gpt4technicalreport} \\
GPT-4.1-mini & \citet{openai2024gpt4technicalreport} \\
\hline
\end{tabular}
\caption{Overview of evaluated large language models.}
\label{tab:models}
\vspace{-15px}
\end{table}

\begin{figure}
    \centering
    \includegraphics[width=0.9\linewidth]{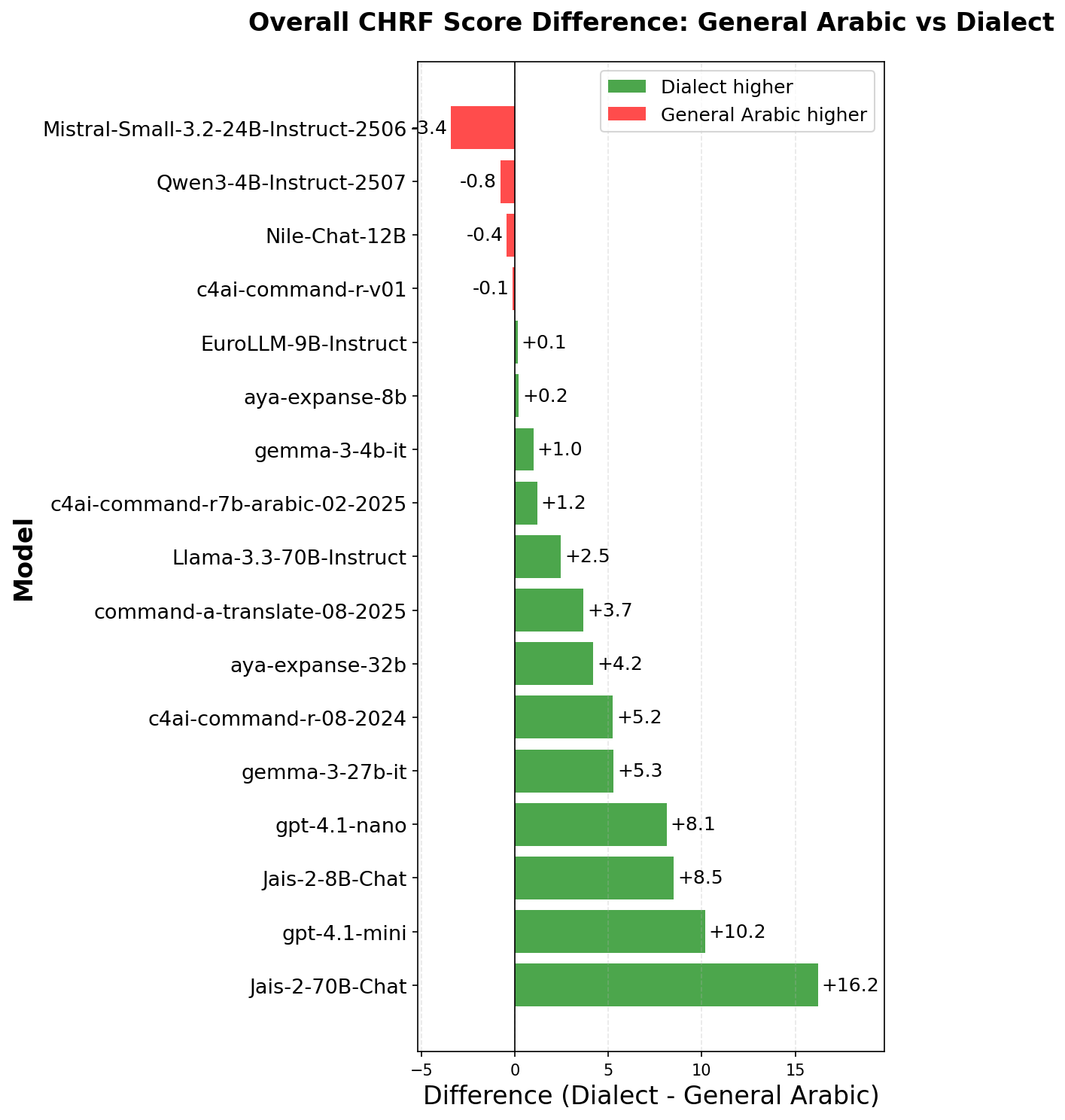}
    \caption{ChrF difference between using dialect-specific vs. general prompt on complete concatenated test set. Green signifies the dialect-specific prompt scored better.}
    \label{fig:dialect_vs_general_prompt}
\end{figure}

\begin{figure*}[tb]
\vspace{-10px}
    \centering
    \includegraphics[width=0.95\linewidth]{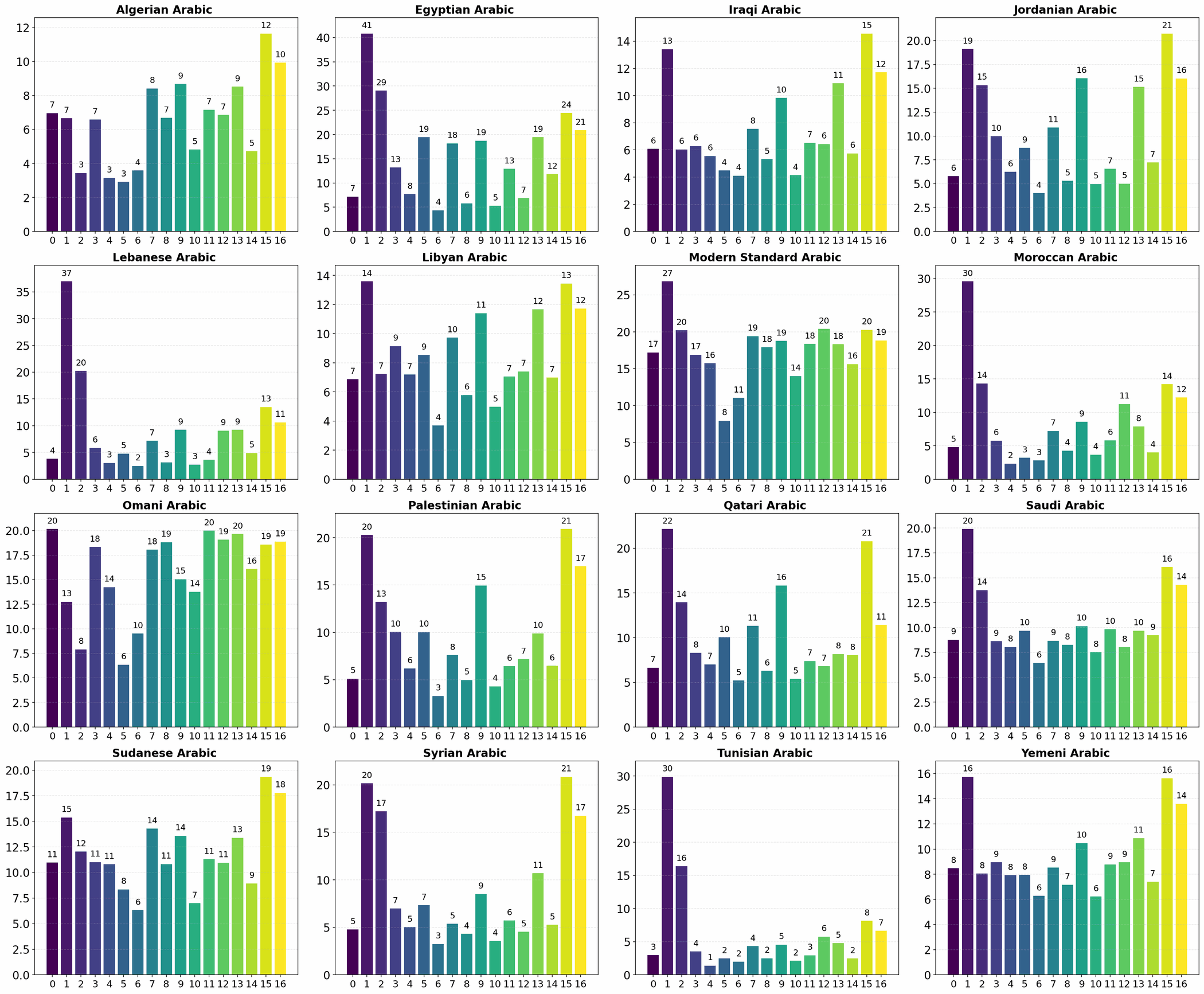}
    \caption{BLEU scores for merged dialect test sets for all models in English to Arabic, using the dialect-specific prompt. The indices on the \textit{x} axis map to the models as follows: 0: EuroLLM-9B-Instruct, 1: Jais-2-70B-Chat, 2: Jais-2-8B-Chat, 3: Llama-3.3-70B-Instruct, 4: Mistral-Small-3.2-24B-Instruct-2506, 5: Nile-Chat-12B, 6: Qwen3-4B-Instruct-2507, 7: aya-expanse-32b, 8: aya-expanse-8b, 9: c4ai-command-r-08-2024, 10: c4ai-command-r-v01, 11: c4ai-command-r7b-arabic-02-2025, 12: command-a-translate-08-2025, 13: gemma-3-27b-it, 14: gemma-3-4b-it, 15: gpt-4.1-mini, 16: gpt-4.1-nano
}
    \label{fig:merged_dialects_bleu}
        \vspace{-15px}
\end{figure*}

\subsection{Dataset Overview}
MADAR  
 is a parallel corpus designed for evaluating machine translation systems across Arabic dialects. The dataset includes 32 test sets covering multiple domains, spanning English and the following dialects: Modern Standard Arabic (MSA),  Moroccan (MA), Tunisian (TN), Algerian (DZ), Libyan (LY), Syrian (SY), Jordanian (JO), Palestinian (PS), Lebanese (LB), Qatari (QA), Omani (OM), Saudi (SA), Yemeni (YE), Iraqi (IQ), Egyptian (EG) and Sudanese (SD). We use the version from AraBench repository \cite{sajjad-etal-2020-arabench}.

\subsection{Evaluated Models}
\label{sec:models}

We evaluate 16 LLMs, shown in Table \ref{tab:models}.

\section{Experimental Settings}
\label{sec:experiments}

\begin{figure*}[tb]
\vspace{-10px}
    \centering
    \includegraphics[width=0.95\linewidth]{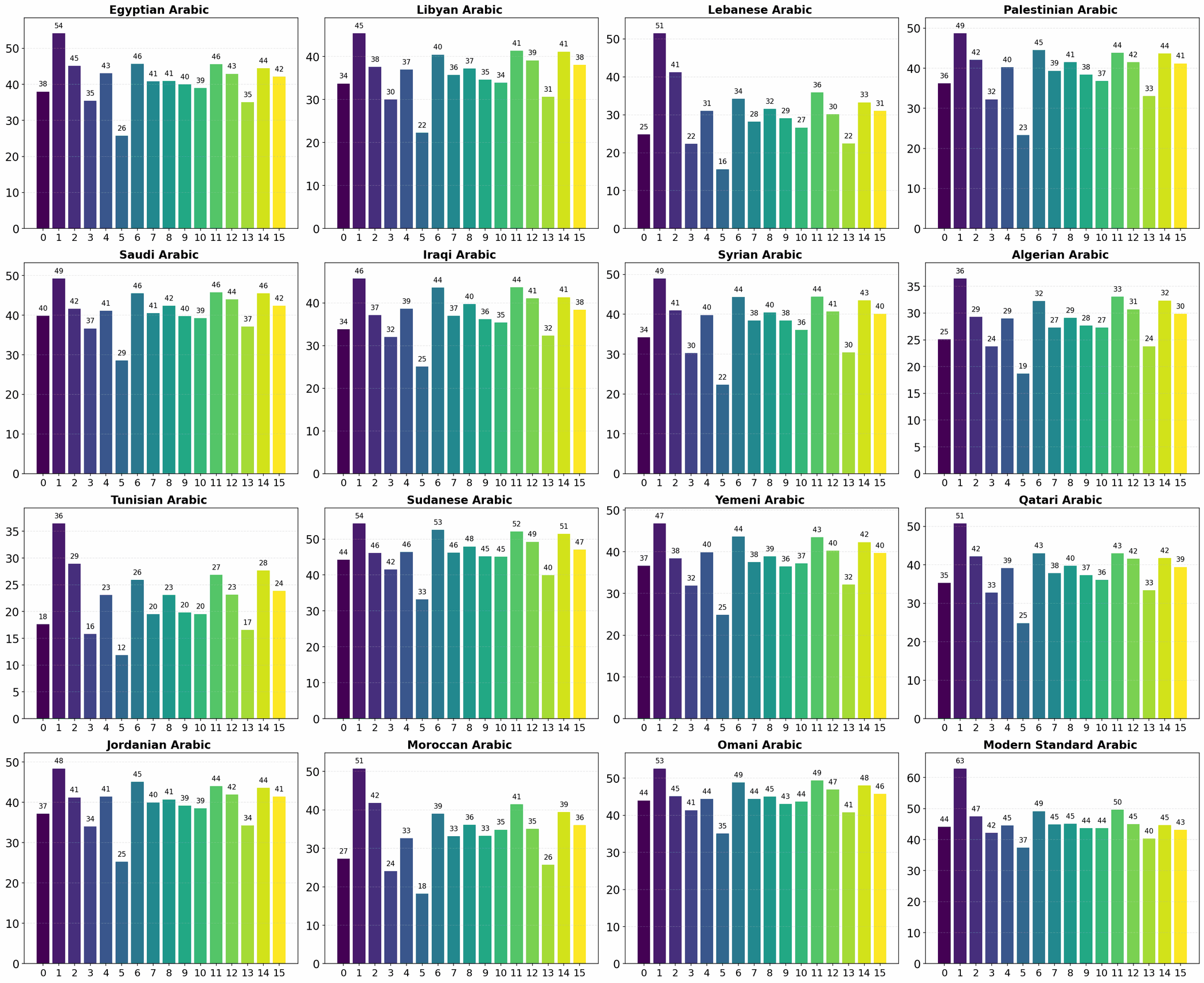}
    \caption{BLEU scores for merged dialect test sets for all models in Arabic to English direction. See Figure \ref{fig:merged_dialects_chrf} for the mapping of the indices on the \textit{x} axis to model names.
}
    \label{fig:merged_dialects_bleu_aren}
    \vspace{-15px}
\end{figure*}

\subsection{Metrics}
 We compute \textbf{BLEU} \citep{papineni-etal-2002-bleu} and \textbf{ChrF} \citep{popovic-2015-chrf} scores using SacreBLEU ~\citep{post-2018-call}. 
 In order to compute \textbf{COMET}~\citep{rei-etal-2020-comet} scores we use the original implementation\footnote{\url{https://github.com/Unbabel/COMET}} and the \texttt{XCOMET-XL} model. 
 \textbf{GEMBA} \cite{kocmi-federmann-2023-large,kocmi-federmann-2023-gemba} is an LLM-as-a-judge approach, where an LLM is prompted to assess translation quality. We use direct assessment without reference, with a score scale from 0 to 100. We instruct the evaluation model to subtract up to 50 points for the wrong dialect used in the translation (see Appendix \ref{sec:prompts} for the exact prompt).
 For \textbf{human evaluation}, we use direct assessment as well, with two separate 0-100 scores for accuracy and dialectalness. 
\subsection{Translation Directions}
We translate the test sets in both directions (English to/from Arabic). For the translation into Arabic, we use \textbf{two different prompts:} \textbf{general Arabic} and \textbf{dialect-specific}. In the dialect-specific, we instruct the model to translate into the dialect specified by the country of the origin of the translation, e.g. \textit{Translate into Egyptian Arabic: {text}}. In the general Arabic prompt, we only instruct the model to translate into Arabic. The exact prompt is shown in Appendix \ref{sec:prompts}. For the Arabic to English direction, we only use one prompt, as we do not specify the source language or dialect.
We compute the scores both on particular files as well as merged datasets, where all the test sets for the given dialect are concatenated.

\subsection{Dialect classification}
We use some of the models (\textit{Jais-2-70B-Chat}, \textit{gemma-3-27b-it}, \textit{aya-expanse-32b} and \textit{Llama-3.3-70B-Instruct}) to classify the dialect of both the produced machine translations and the reference human translations. While GPT models showed a promising performance across tasks, we omit them from these experiments due to budget constraints. We aim to assess if these LLMs can identify the dialects (by classifying the reference translations), and, provided they can, we use them to evaluate dialectalness of the produced machine translations. The classification prompt is shown in Appendix \ref{sec:prompts}.
\subsection{Inference Setup}
The OpenAI models are accessed through OpenAI API with default parameters. We deploy other models using VLLM with default decoding parameters.

\subsection{Postprocessing}
We have noticed that some LLMs produce repetitions of a single token for some of the inputs. In case the produced output is more than 5 times longer than the source, either in tokens or in characters, we crop the output to the input token length.

We use CAMeL Tools \cite{obeid-etal-2020-camel} to normalize both the references and the translations. We use Alef Maksura, Alef, Teh Marbuta and Hamza normalizations and we also convert Arabic numerals to Latin ones. \XXX{@Rawan: is this the correct way to formulate this?}

\section{Results and Analysis}
\label{sec:results}


In Figure \ref{fig:dialect_vs_general_prompt}, we compare system-level (i.e., on concatenation of all test sets) ChrF scores for general vs. dialect-specific prompts across models. The figure demonstrates which models react to dialect-specific prompts and to what extent. Many of the models are able to distinguish between dialects and produce more appropriate translations based on the dialect specification, resulting in higher overall ChrF scores when the target dialect is specified.

In Figure \ref{fig:merged_dialects_bleu}  we present the English to Arabic dialect-level (i.e. on concatenation of test sets pertaining to one dialect) BLEU scores, for the dialect-specific prompt for all models. ChrF scores for the same configuration are located in Appendix \ref{sec:appendix_results}, Figure \ref{fig:merged_dialects_chrf}. We see that for many models,  BLEU scores under 10, which usually signal unusable translation, are common. Notable exceptions are the Jais-2 and gpt-4.1 models, which result in at least moderate scores for most of the dialects, except the Algerian, Iraqi, Yemeni and Libyan dialects. Even for these models however, many of the dialect BLEU scores are not higher than 20, indicating that the translations are not very similar to the reference. To gauge whether the low scores are indicative of poor translation quality, or are more caused by the orthographic and morphological richness and diversity of the dialects, we carry out a small manual evaluation for the Egyptian dialect, described later.

\begin{figure*}[tb]
    \centering
    \includegraphics[width=1\linewidth]{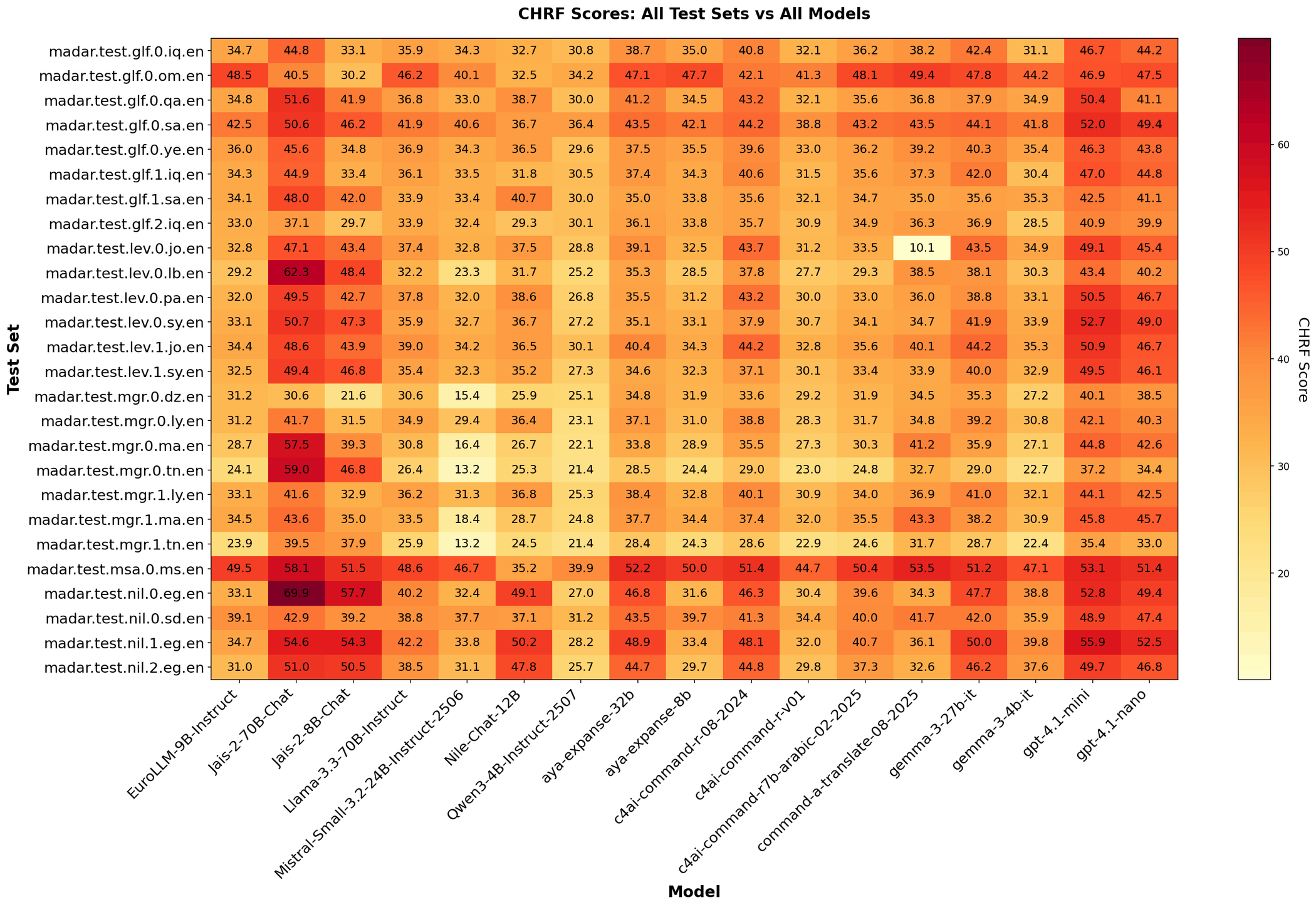}
    \caption{ChrF scores for individual test sets for all models in English to Arabic direction. Translation produced using dialect-specific prompt.}
    \label{fig:heatmap_chrf}
        \vspace{-15px}
\end{figure*}

\begin{table}
\small
\centering
  \begin{adjustbox}{max width=\linewidth}
\begin{tabular}{l@{\hspace{3pt}}l@{\hspace{3pt}}c@{\hspace{3pt}}c@{\hspace{3pt}}c}
\toprule
Model & Size & Wins & Avg. Rank \\
\midrule
gpt-4.1-mini & N/A & 16 & 2.1 \\
Jais-2-70B-Chat & 70B & 14 & 2.9 \\
gpt-4.1-nano & N/A & 0 & 3.5 \\
gemma-3-27b-it & 27B & 0 & 5.3 \\
c4ai-command-r-08-2024 & 32B & 0 & 6.0 \\
aya-expanse-32b & 32B & 0 & 6.7 \\
command-a-translate-08-2025 & 111B & 2 & 7.1 \\
Jais-2-8B-Chat & 8B & 0 & 7.8 \\
Llama-3.3-70B-Instruct & 70B & 0 & 9.6 \\
c4ai-command-r7b-arabic-02-2025 & 7B & 0 & 9.6 \\
Nile-Chat-12B & 12B & 0 & 10.8 \\
EuroLLM-9B-Instruct & 9B & 0 & 11.1 \\
aya-expanse-8b & 8B & 0 & 11.7 \\
gemma-3-4b-it & 4B & 0 & 12.7 \\
Mistral-Small-3.2-24B-Instruct-2506 & 24B & 0 & 14.8 \\
c4ai-command-r-v01 & 35B & 0 & 15.0 \\
Qwen3-4B-Instruct-2507 & 4B & 0 & 16.4 \\
\bottomrule
\end{tabular}
\end{adjustbox}
\caption{Average rankings of all models based on ChrF scores using dialect-specific prompts. }
\label{tab:ranking}
    \vspace{-15px}
\end{table}

We show the ChrF scores on individual test sets in Figure \ref{fig:heatmap_chrf}. We see that \texttt{gpt-4.1-mini} and \texttt{Jais-2-70B-Chat} models have the highest scores across all test sets, with a very few exceptions. This is also demonstrated by the ChrF-based ranking of each model, computed across all test sets, which is shown in Table \ref{tab:ranking}. Across all models, we see high scores on the MSA test set, showing that most of the models possess good capabilities for translation into MSA, but only some of them are capable of producing dialectal Arabic translations.

The BLEU scores for the reverse direction, Arabic-to-English, are presented in Figure \ref{fig:merged_dialects_bleu_aren}. The differences between models are less pronounced here and the overall scores are higher as most models can produce high-quality English translations for a dialectal Arabic source text. \texttt{Jais-2-70B-Chat} scores the best on all test sets. 

Similar plots for the COMET-XL scores are shown in Appendix \ref{sec:appendix_results}, in Figures \ref{fig:comet_refless_dialect} and \ref{fig:comet_ref_dialect} for the reference-based and reference-less versions. These results demonstrate that COMET is not suitable for our use-case, which is caused by the fact that these models do not allow the user to specify source and target languages. This could be partially mitigated by using the reference-based version, but even then, the model largely prefers the MSA translations over the dialectal ones. We make this observation as an argument for including language pair specification into future automatic MT metrics. 

Other results are located in Appendix \ref{sec:appendix_results}.
\begin{table}
\centering
\small

\begin{tabular}{l@{\hspace{3pt}}rr@{\hspace{3pt}}rr}
\toprule
Dialect & \multicolumn{2}{c@{\hspace{3pt}}}{BLEU} & \multicolumn{2}{c@{\hspace{3pt}}}{CHRF} \\\cmidrule(lr){2-3} \cmidrule(lr){4-5} & General & Dialect & General & Dialect \\
\midrule
Algerian  & 7.0 & 11.6 & 32.8 & 40.1 \\
Egyptian  & 6.0 & 23.6 & 32.5 & 52.2 \\
Iraqi  & 5.9 & 14.6 & 35.8 & 44.8 \\
Jordanian  & 5.7 & 20.7 & 34.8 & 50.0 \\
Lebanese  & 3.3 & 14.0 & 30.5 & 44.4 \\
Libyan  & 6.4 & 13.4 & 33.3 & 43.1 \\
MSA  & 16.6 & 16.4 & 48.5 & 48.2 \\
Moroccan  & 3.2 & 10.6 & 27.7 & 41.2 \\
Omani  & 21.0 & 18.6 & 50.7 & 46.9 \\
Palestinian  & 5.4 & 20.9 & 33.1 & 50.5 \\
Qatari  & 6.6 & 14.0 & 35.8 & 43.9 \\
Saudi  & 9.5 & 16.1 & 40.3 & 47.3 \\
Sudanese  & 11.1 & 19.3 & 41.0 & 48.9 \\
Syrian  & 4.8 & 20.8 & 34.3 & 51.1 \\
Tunisian  & 2.6 & 7.6 & 25.3 & 36.1 \\
Yemeni  & 8.5 & 15.6 & 37.4 & 46.3 \\
\bottomrule
\end{tabular}
\caption{Average merged dialect scores for GPT-4.1-mini: General vs. dialect-specific prompt}
\label{tab:dialect_prompt_comparison}
    \vspace{-15px}
\end{table}

\XXX{@Rawan: Why is Omani better without dialect-specific prompt?}

\subsection{GEMBA}
\begin{figure}[h]
    \centering
    \includegraphics[width=1\linewidth]{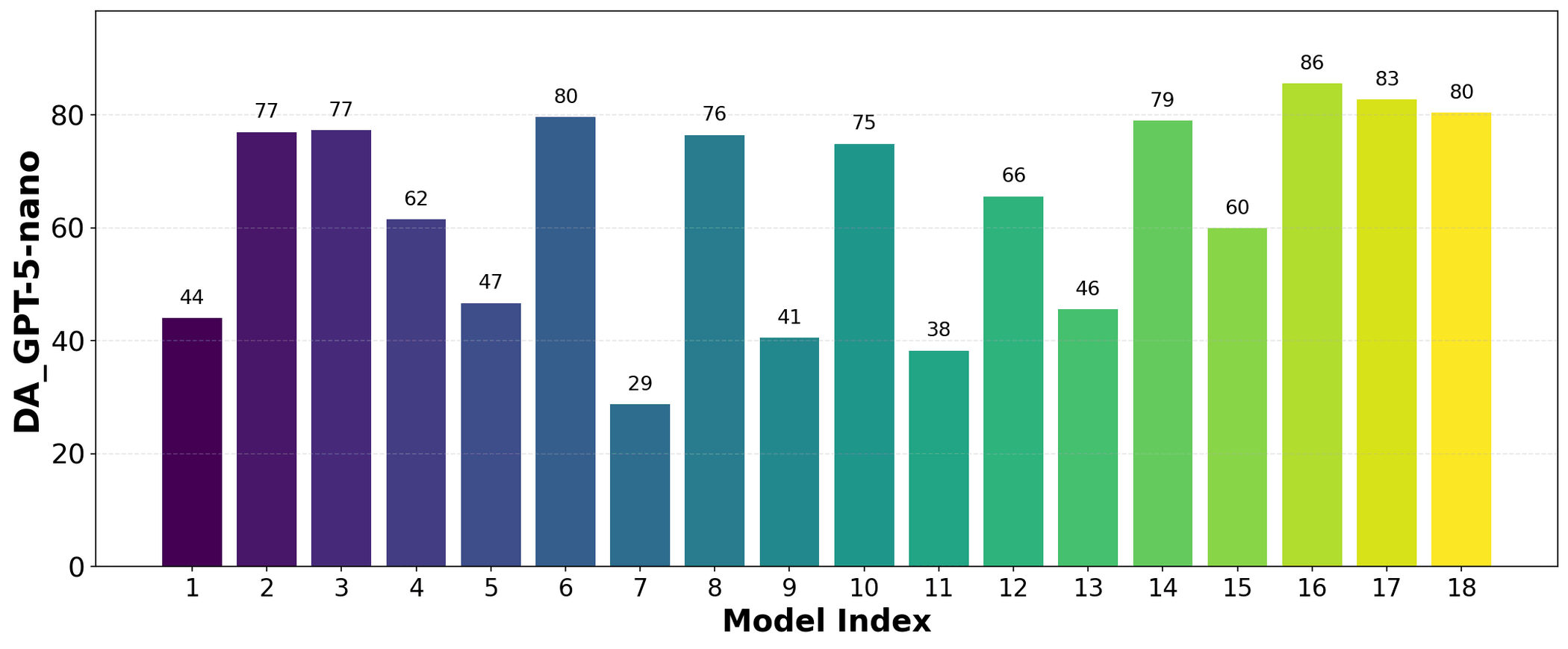}
    \caption{GEMBA GPT-5-nano DA evaluation scores for all models on madar.test.nil.1.eg, using the dialect-specific prompt. See Figure  \ref{fig:merged_dialects_bleu} for the mapping of the indices on the \textit{x} axis to model names.
}
    \label{fig:egypt_gemba}
        \vspace{-15px}
\end{figure}
We also used an LLM-as-a-judge approach (GEMBA) to score the translations. We simulated a direct assessment setting, with a 0-100 scale, up 50 points being awarded for the translation accuracy and 50 points being reserved for the dialect appropriateness. We used \texttt{gpt-5-nano} as the judge. Due to budget constraints, we have only run the scoring on the same test set we have performed our human evaluation on, \texttt{madar.test.nil.1.eg}. The results are shown in Figure \ref{fig:egypt_gemba}.  There are multiple well-performing models, and the results among the top models are more even than according to ChrF and BLEU scores. \XXX{@Rawan: Is something like this true? Even though we normalize the translations and the references? Based on manual analysis, we hypothesize that the large BLEU score gap is caused by the fact that Jais-2 model might have been trained on data more closely resembling the test set references, for example, in terms of orthography. Other models that score well according to gpt-5-nano, but not according to BLEU or ChrF, produce valid dialectal translations, but in a different ortographic standard, which is only partially mitigated by the normalization.I think it is not orthography but based on the manual evaluation it is the phrasing rather than orthographic standard @Josef I think this might have to do with the phrading more than orthograhpy, for example Kowyyis and Tamam both mean fine in Egyptian arabic, Both of them will have similar score in human evaluation but vastly different BLEU/ChrF scores}

\begin{figure}[h]
    \centering
    \includegraphics[width=1\linewidth]{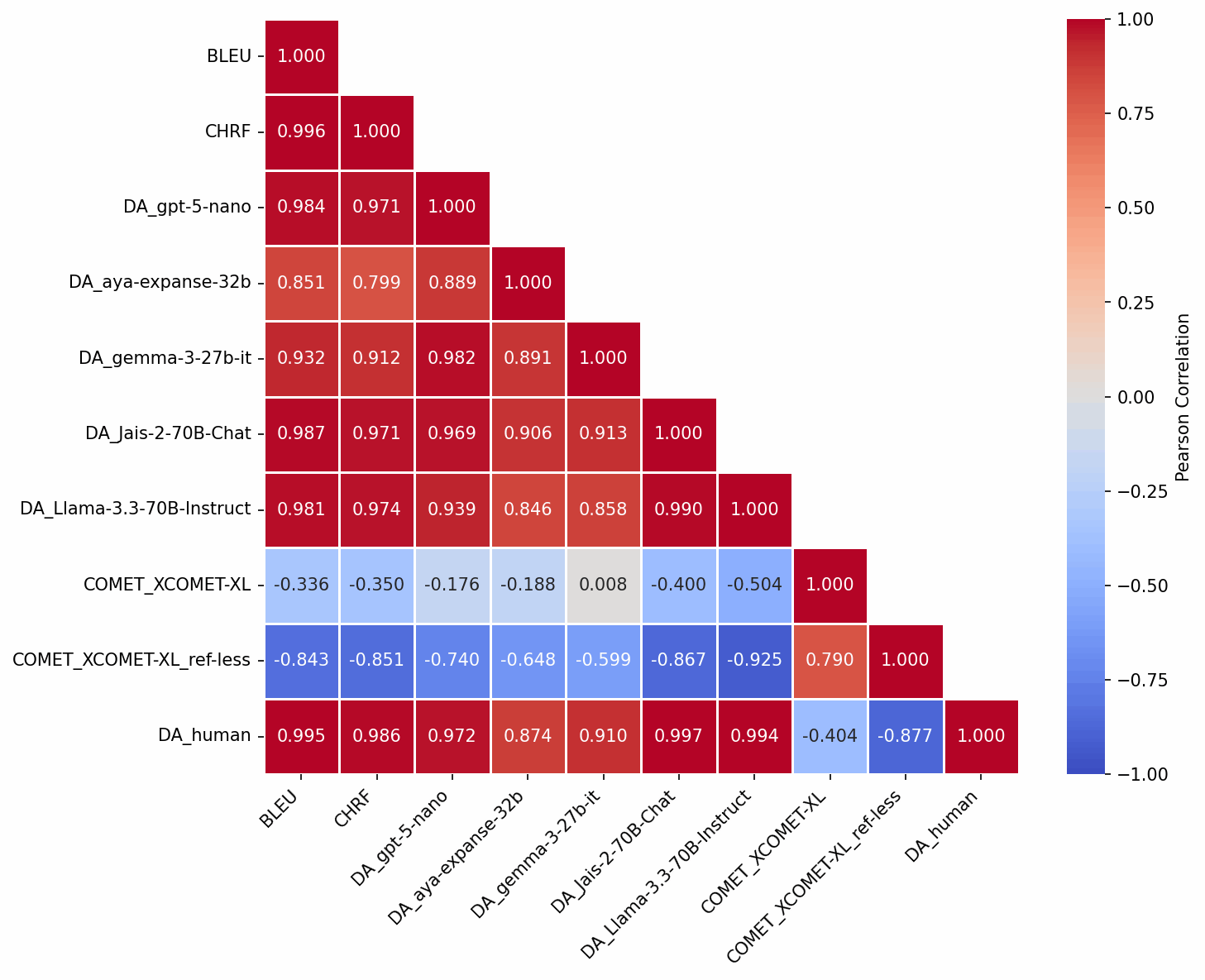}
    \caption{Pearson's correlation of system-level automatic metrics and human DA on the first 100 sentences of \texttt{madar.test.nil.1.eg} for translation created with the dialect-specific prompt
}
    \label{fig:correlation}
\vspace{-15px}
\end{figure}

\begin{figure*}[tb]
    \centering
    \includegraphics[width=1\linewidth]{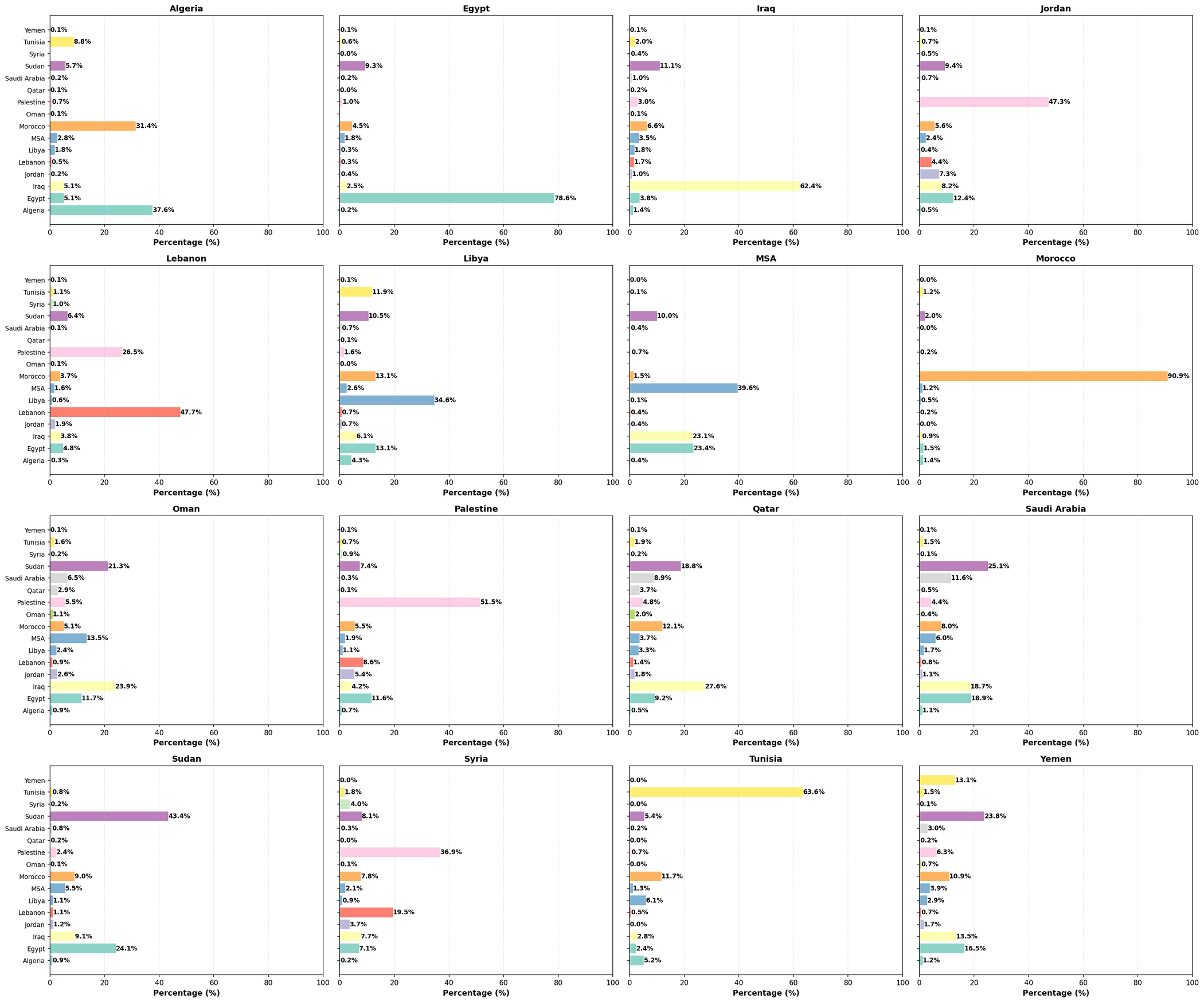}
    \caption{Distribution of dialects in reference translations. The dialects are classified on the sentence level, using \texttt{Jais-2-70B-Chat}.}
    \label{fig:dialect_ref_by_jais}
        \vspace{-15px}
\end{figure*}

\subsection{Manual analysis}\

The manual error analysis followed a two-dimensional annotation scheme consisting of translation \textbf{accuracy} and \textbf{dialectness}, which were assessed independently for each generated output.

Accuracy annotations capture semantic adequacy errors, indicating whether the meaning of the English source sentence is correctly conveyed in the translation. Errors under this dimension include omissions, additions, mistranslations, and incorrect semantic relations.

Dialectness annotations target the dialectal realization, reflecting the extent to which the generated translation conforms to the lexical, morphological, and syntactic conventions of the target dialect. Errors in this category include the use of MSA or other non-target dialect forms, unnatural code-switching, and dialect-inappropriate constructions.

Crucially, the two annotation dimensions are orthogonal: a translation may be semantically accurate while failing to realize the target dialect, or conversely, may exhibit strong dialectal features despite conveying an incorrect or incomplete meaning.  
We performed a small-scale manual analysis on a sample of 100 sentences from \texttt{madar.test.nil.1.eg} test set. \texttt{Jais-2-70B-Chat} and \texttt{gpt-4.1-mini} were selected as the two best-performing systems according to automatic evaluation scores, and \texttt{EuroLLM-9B-Instruct} and \texttt{C4ai-command-r-08-2024}, which achieved comparatively lower automatic scores. Including both high- and lower-performing models allowed us to examine qualitative differences in error patterns and dialectal realization across performance tiers. 
Figure \ref{fig:correlation} shows Pearson's system-level correlation between automatic metrics and human assessment (the mean of dialectalness and accuracy DA scores). The scores suggest the suitability of BLEU, ChrF and LLM-produced DA scoring, which has the advantage of not needing a reference translation. It also further demonstrantes the unsuitability of COMET scores for our use-case, already discussed earlier. However, due to the small-scale nature of our human evaluation, further data is needed to confirm these observations.
\XXX{Examples, probably better to show them in the appendix}

\begin{table}[!htp]\centering
\small
\begin{tabular}{lrrrrr}\toprule
\textbf{Model} &\textbf{J70B} &\textbf{G-mini} &\textbf{ELLM-9B} &\textbf{CR-824} \\\midrule
\textbf{Accuracy} &94.6 &96.3 &93.0 &93.4 \\
\textbf{Dialect} &98.5 &96.0 &36.4 &86.8 \\
\textbf{Mean Score} &96.5 &96.2 &64.7 &90.1 \\
\bottomrule
\end{tabular}
\small
\caption{Manual evaluation scores for accuracy and dialectalness on the translation of the first 100 sentences of \texttt{madar.test.nil.1.eg} produced with the dialect-specific prompt by the following models: Jais-2-70B-Chat (J70B), GPT-4.1-mini (G-mini), EuroLLM-9B-Instruct (ELLM-9B), Command-r-08-2024 (CR-824). }\label{tab:manual_eval}
\end{table}

\subsection{Dialect classification}

We employed LLMs to identify the dialect of a text. First, we classified the reference translations to verify the ability of the models to correctly assess the dialect used in a text. We assume that in a test set for a given dialect, a large part of sentences posses the features of that dialect. We show the classification results produced by the \texttt{Jais-2-70B-Chat} model for the reference human translations in Figure \ref{fig:dialect_ref_by_jais}. \XXX{TODO: discuss the results, similarity of dialects, e.g. is Sudan vs. Egypt considered misclassification? } Plots for translations produced by the LLMs are shown in Appendix \ref{sec:app_dial}. We observe that in general, the dialectical human translation exhibit higher percentage of sentences classified as belonging to the given dialect, or a related one (e.g. Egypt and Sudan in the Sudan test set).
\section{Discussion}
\label{sec:discussion}

When considered jointly, automatic and manual evaluations reveal complementary strengths and limitations of current LLMs for English-to-Dialectal Arabic MT. Models trained with substantial dialectal Arabic data, such as Jais-2-70B-Chat, outperform general-purpose systems, particularly in producing outputs with high dialectness. By contrast, models optimized for broad multilingual coverage, such as EuroLLM-9B-Instruct, tend to generate more standardized or mixed varieties. 

Although automatic metrics like BLEU and ChrF correlate well with human judgments overall on our small sample, manual evaluation provides finer-grained insights into dialectal realization. 

Human evaluation further reveals a distinct pattern in the \texttt{c4ai-command-r-08-2024} model’s handling of gender-ambiguous inputs: unlike other models, which default to masculine morphology, it frequently resolves ambiguity using feminine forms. This behavior suggests an implicit mitigation strategy that is not captured by automatic metrics and raises important questions about bias and sociolinguistic alignment in dialectal MT.

\XXX{@Rawan: Maybe something about our prompting, noting that using just a country name to specify a dialect is not optimal since there can be multiple dialects in one country, or one dialect spanning multiple countries? yes I think we could add that}

\section{Conclusion}
\label{sec:conclusion} 
We presented a large-scale evaluation of 16 LLMs on English to Arabic MT across 16 Arabic dialects, combining BLEU/ChrF/COMET, LLM-as-a-judge and a human assessment study. Across metrics and test sets, performance is highly uneven: most models translate well into MSA, but many struggle to reliably produce the requested dialect, with particularly low scores common for several dialects.

Arabic-specialized and strong commercial models are the most consistent for dialectal generation. Dialect-specific prompting generally improves dialectal outputs substantially, confirming that some LLMs can condition on dialect instructions. GPT-4.1-mini and Jais-2-70B dominate system-level rankings, while several other multilingual models frequently default to MSA, even when prompted for a dialectal translation. 

Our analyses also show limitations of current automatic MT evaluation for dialectal Arabic. Reference-based metrics can substantially undervalue valid dialectal variants due to non-standard orthography and high surface-form variability; this is supported by our manual evaluation. COMET score proved poorly suited to this setting because it cannot be constrained to penalize “wrong-dialect” outputs and tends to favor MSA-like texts.

Our comparison is intended to support researchers and practitioners in selecting suitable models for deployment in machine translation applications. We acknowledge that the test set may have been included in the models’ training data, and therefore, the high scores should be interpreted with caution. Nevertheless, low scores are still informative, as they indicate models that are unable to generate the target dialects, and this insight remains practically useful.


\section*{Limitations}
The main limitation of our work is the fact that for many of the models, we do not have access to the training data. It is possible that the MADAR test set we use for evaluation was seen by the models. Another limitation is that we did not optimize the translation prompt for specific models, which could lead to a lower translation quality for some of the systems. We also note that using a country name as a specification of a dialect is not optimal, as some dialects span multiple countries, or there are multiple dialects used within a single country. Another limitation is that we performed only a small-scale human evaluation and mostly rely on automated metrics, which have many known issues.

\section*{Acknowledgments}
This work was partially supported by SVV project number 260 821, by Czech Ministry of Education, Youth and Sports (grant MŠMT OP JAK Mezisektorová spolupráce CZ.02.01.01/00/23\_020/0008518) and by EU EDF project ALADAN,
Grant Agreement No 101102545. 

It has been using language resources and tools developed and/or stored and/or distributed by the LINDAT/CLARIAH-CZ project of the Ministry of Education, Youth and Sports of the Czech Republic (project LM2023062).
\bibliography{latex/custom,latex/anthology_0,latex/anthology_1}

\appendix

\section{Additional results}
\label{sec:appendix_results}
\subsection{English to Arabic}
\begin{figure*}[tb]
    \centering
    \includegraphics[width=1\linewidth]{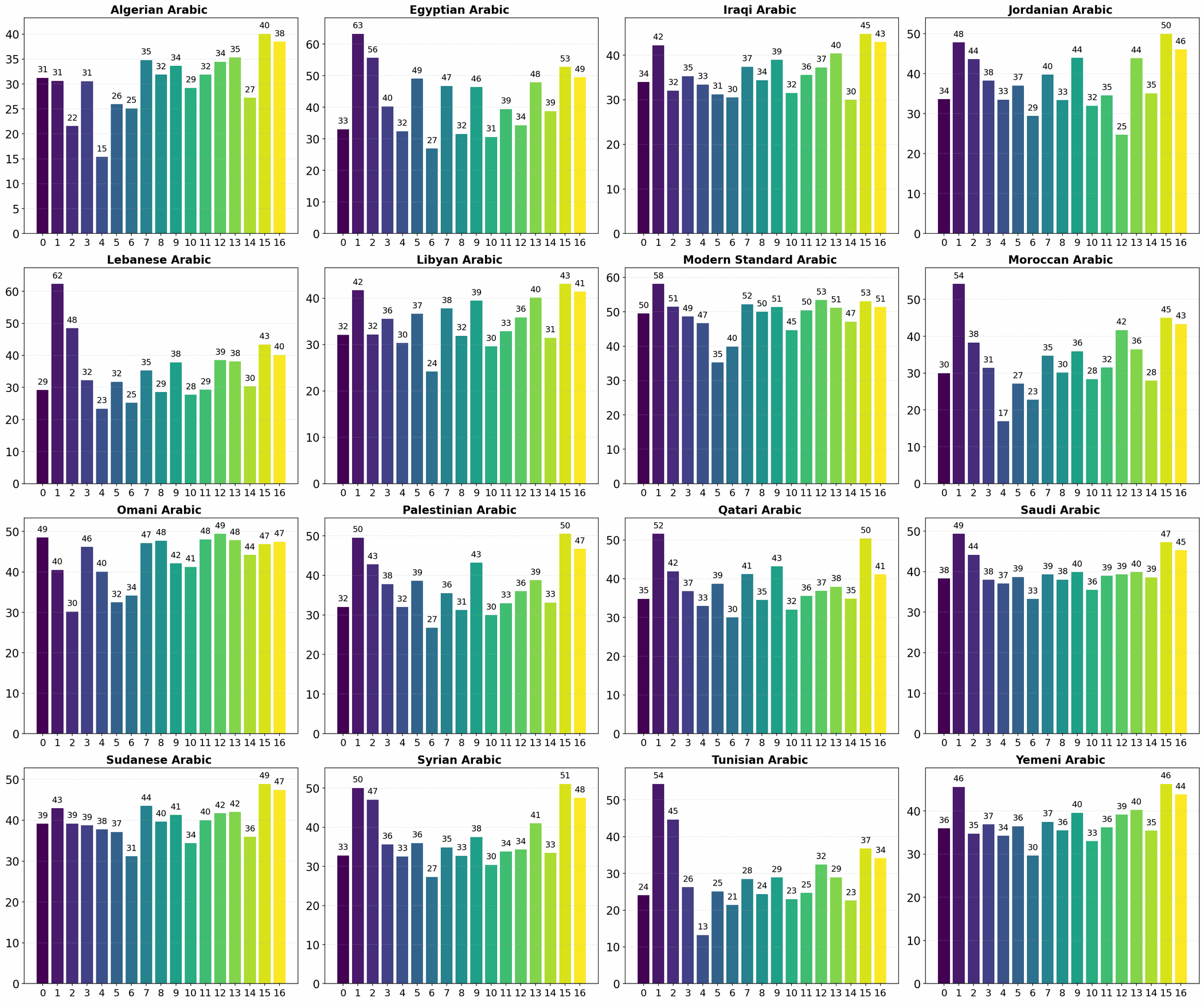}
    \caption{ChrF scores for merged dialect test sets for all models in English to Arabic, using the dialect-specific prompt.  See Figure \ref{fig:merged_dialects_bleu} for the mapping of the indices on the \textit{x} axis to model names. 
}
    \label{fig:merged_dialects_chrf}
\end{figure*}

\begin{table}[htbp]
\small
\centering

\begin{tabular}{l@{\hspace{3pt}}r@{\hspace{3pt}}r@{\hspace{3pt}}r}
\toprule
Model & Size & BLEU & CHRF \\
\midrule
\multicolumn{4}{l}{\textit{Arabic-Specialized}} \\
\midrule
Jais-2-70B-Chat & 70B & 25.4 & 51.8 \\
Jais-2-8B-Chat & 8B & 14.5 & 41.6 \\
Nile-Chat-12B & 12B & 7.3 & 34.1 \\
c4ai-command-r7b-arabic-02-2025 & 7B & 8.1 & 34.5 \\
\midrule
\multicolumn{4}{l}{\textit{Multilingual}} \\
\midrule
EuroLLM-9B-Instruct & 9B & 7.0 & 33.3 \\
Llama-3.3-70B-Instruct & 70B & 8.6 & 35.3 \\
Mistral-Small-3.2-24B-Instruct-2506 & 24B & 6.0 & 28.4 \\
Qwen3-4B-Instruct-2507 & 4B & 4.6 & 27.8 \\
aya-expanse-32b & 32B & 9.9 & 38.0 \\
aya-expanse-8b & 8B & 6.5 & 33.2 \\
c4ai-command-r-08-2024 & 32B & 11.1 & 39.1 \\
c4ai-command-r-v01 & 35B & 5.4 & 30.2 \\
command-a-translate-08-2025 & 111B & 8.7 & 37.7 \\
gemma-3-27b-it & 27B & 10.5 & 39.2 \\
gemma-3-4b-it & 4B & 7.1 & 32.7 \\
\midrule
\multicolumn{4}{l}{\textit{Commercial API}} \\
\midrule
gpt-4.1-mini & N/A & 15.0 & 44.8 \\
gpt-4.1-nano & N/A & 12.4 & 42.1 \\
\midrule
\bottomrule
\end{tabular}
\caption{Overall BLEU and ChrF scores computed on the concatenation of all test sets translated using the dialect-specific prompts. }
\label{tab:overall_scores_forward_dialect}
\end{table}

\begin{figure*}[tb]
    \centering
    \includegraphics[width=1\linewidth]{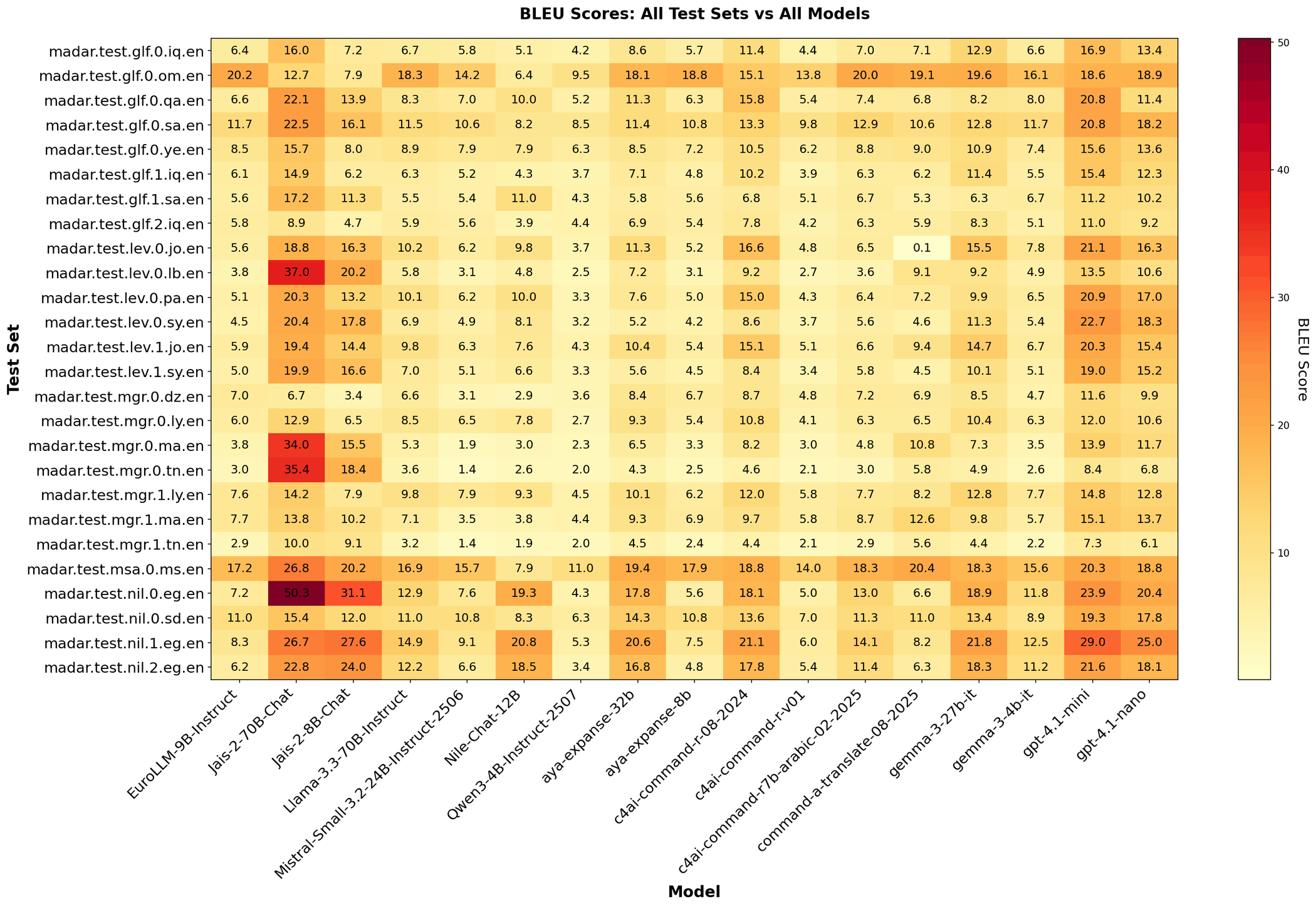}
    \caption{BLEU scores for individual test sets for all models in English to Arabic direction. Translation produced using a dialect-specific prompt.}
    \label{fig:heatmap_bleu}
\end{figure*}

\begin{figure*}[tb]
    \centering
    \includegraphics[width=1\linewidth]{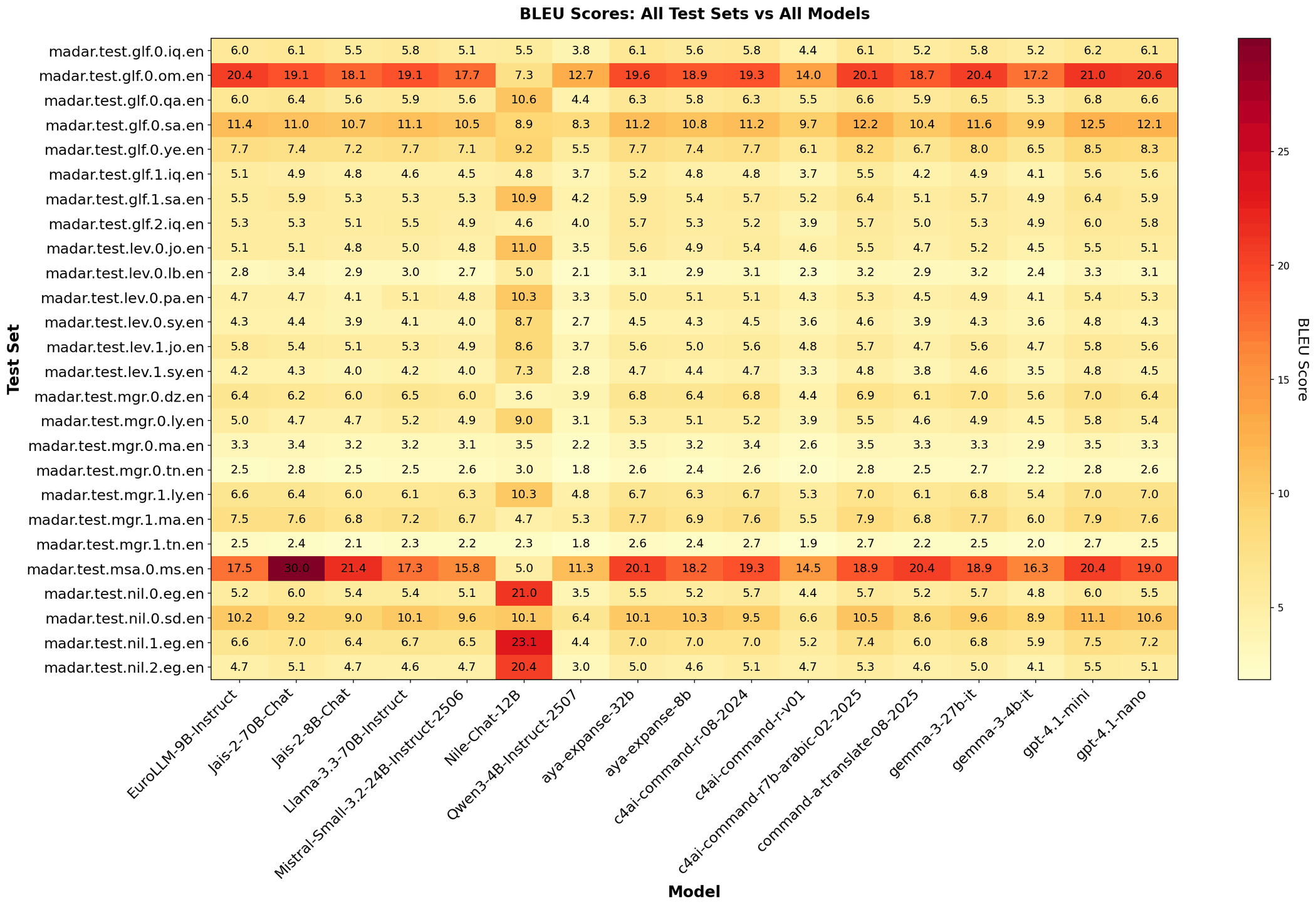}
    \caption{BLEU scores for individual test sets for all models in English to Arabic direction. Translation produced using a general Arabic prompt.}
    \label{fig:heatmap_bleu_arabic_general}
\end{figure*}

\begin{figure*}[tb]
    \centering
    \includegraphics[width=1\linewidth]{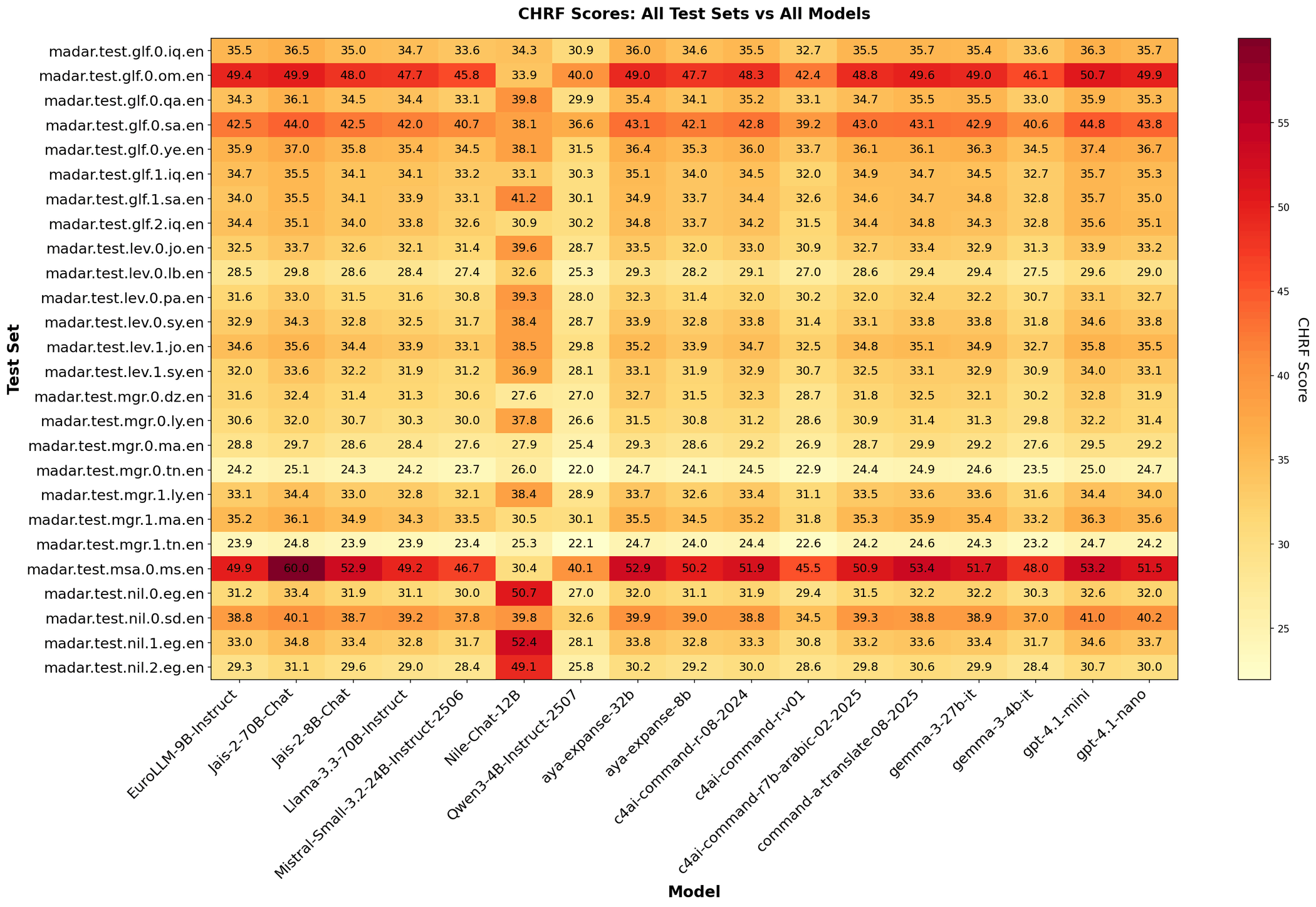}
    \caption{ChrF scores for individual test sets for all models in English to Arabic direction. Translation produced using a general Arabic prompt.}
    \label{fig:heatmap_chrf_arabic_general}
\end{figure*}

\begin{figure*}[tb]
    \centering
    \includegraphics[width=1\linewidth]{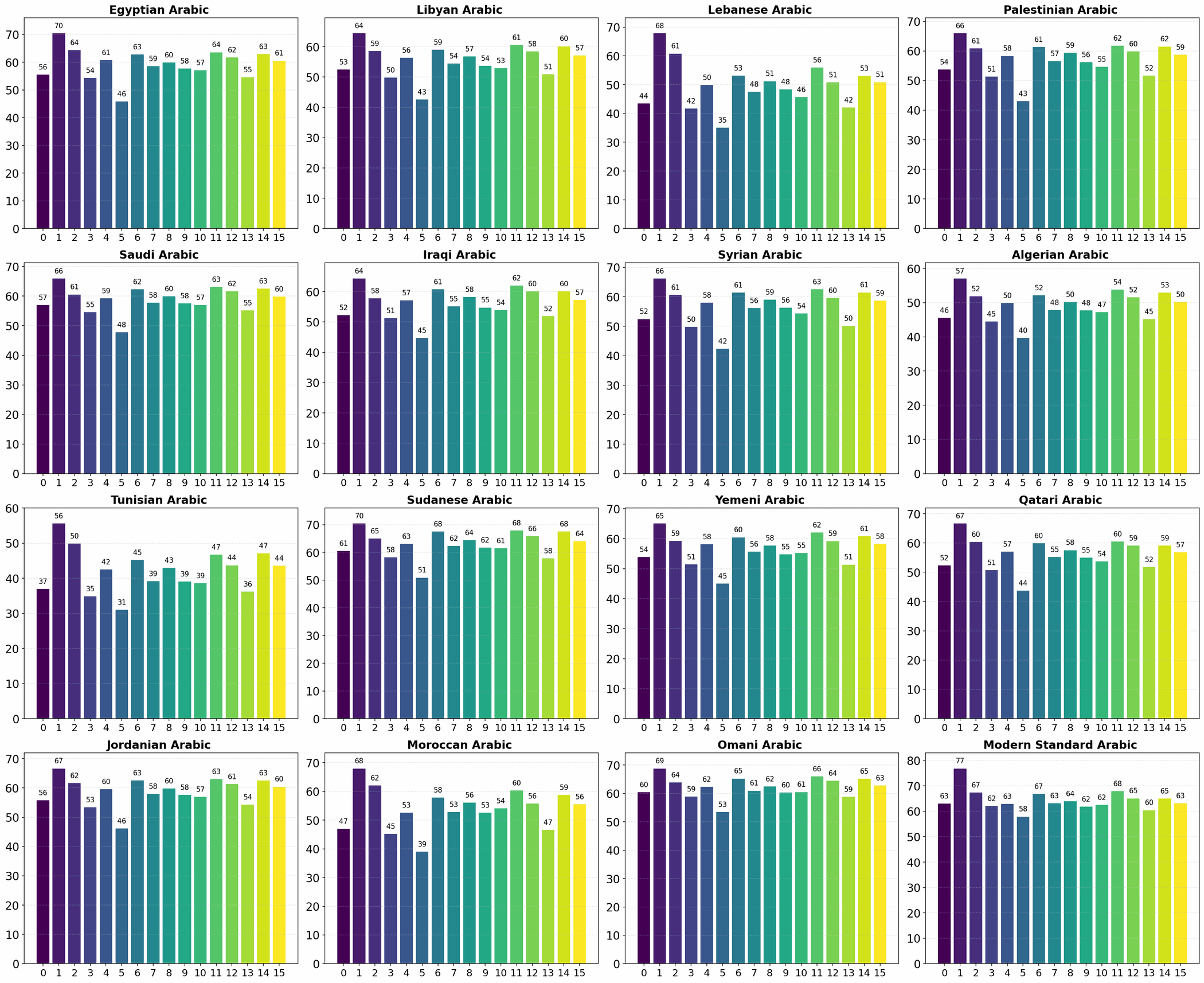}
    \caption{ChrF scores for merged dialect test sets for all models in Arabic to English direction. See Figure \ref{fig:merged_dialects_bleu} for the mapping of the indices on the \textit{x} axis to model names.
}
    \label{fig:merged_dialects_chrf_aren}
\end{figure*}

The total BLEU and ChrF scores, computed on the concatenation of all test sets, are presented in Table \ref{tab:overall_scores_forward_dialect}. These scores again show the leading performance of Jais-2 and gpt-4.1 models, followed by Command-R and Gemma 3. 

Figure \ref{fig:heatmap_bleu} shows the BLEU scores on single test sets for translations produced using the dialect-specific prompt. For comparison,  Figure \ref{fig:heatmap_bleu_arabic_general} shows BLEU scores for translations produced with general Arabic prompt. Figure \ref{fig:heatmap_chrf_arabic_general} shows the ChrF scores for translations with the general prompt and can be compared to the dialect-specific prompt scores in Figure \ref{fig:heatmap_chrf}. Comparison of plots allows to distinguish which models are sensitive to dialect-specific prompts and able to produce adequate dialectal realization of the translation.

\subsection{Arabic to English}

Figure \ref{fig:merged_dialects_chrf_aren} shows ChrF scores for Arabic to English translations. The scores are comparatively higher compared to English to Arabic scores (Figure \ref{fig:merged_dialects_chrf}), which may reflect both better translation quality and lower morphological complexity of English.
\subsection{COMET scores}

\begin{figure*}[tb]
    \centering
    \includegraphics[width=1\linewidth]{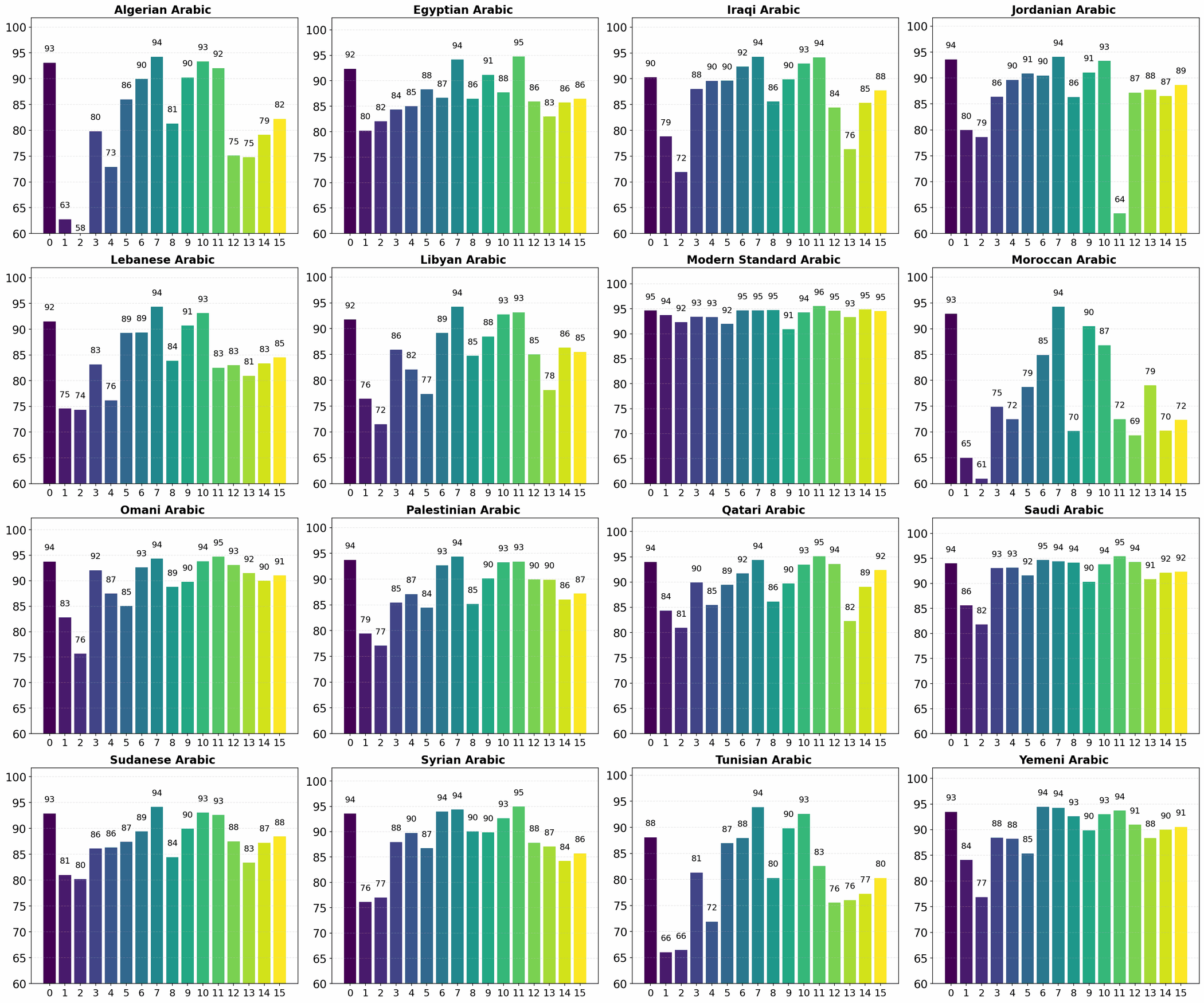}
    \caption{COMET scores \textbf{without} using the reference for merged dialect test sets for all models in English to Arabic, using the \textbf{dialect-specific prompt}.  See Figure \ref{fig:merged_dialects_bleu} for the mapping of the indices on the \textit{x} axis to model names. 
}
    \label{fig:comet_refless_dialect}
\end{figure*}

\begin{figure*}[tb]
    \centering
    \includegraphics[width=1\linewidth]{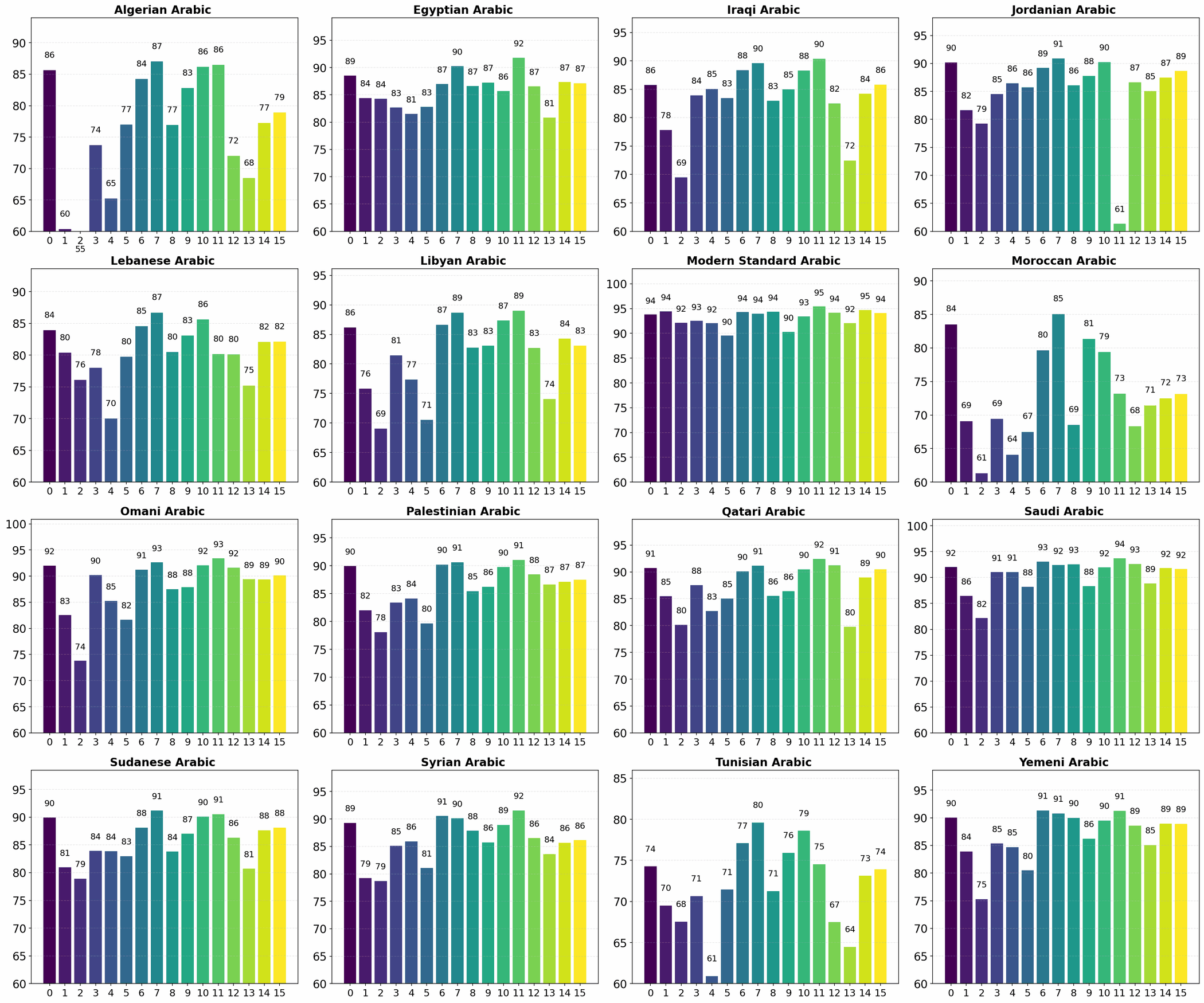}
    \caption{COMET scores \textbf{using the reference} for merged dialect test sets for all models in English to Arabic, using the \textbf{dialect-specific prompt}.  See Figure \ref{fig:merged_dialects_bleu} for the mapping of the indices on the \textit{x} axis to model names. 
}
    \label{fig:comet_ref_dialect}
\end{figure*}

\begin{figure*}[tb]
    \centering
    \includegraphics[width=1\linewidth]{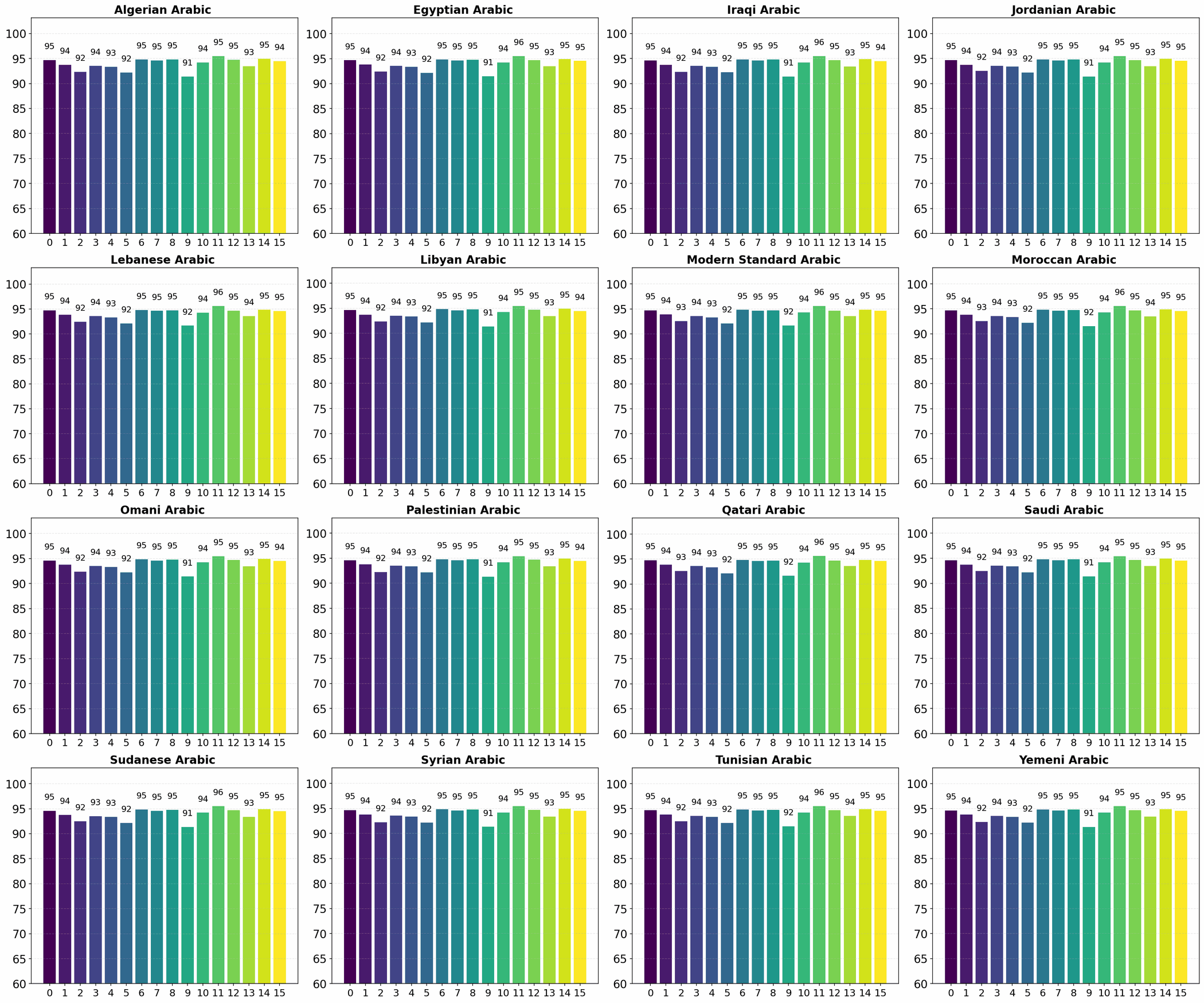}
    \caption{COMET scores without using the reference for merged dialect test sets for all models in English to Arabic, using the  \textbf{general Arabic prompt}  See Figure \ref{fig:merged_dialects_bleu} for the mapping of the indices on the \textit{x} axis to model names. 
}
    \label{fig:comet_refless_general}
\end{figure*}

\begin{figure*}[tb]
    \centering
    \includegraphics[width=1\linewidth]{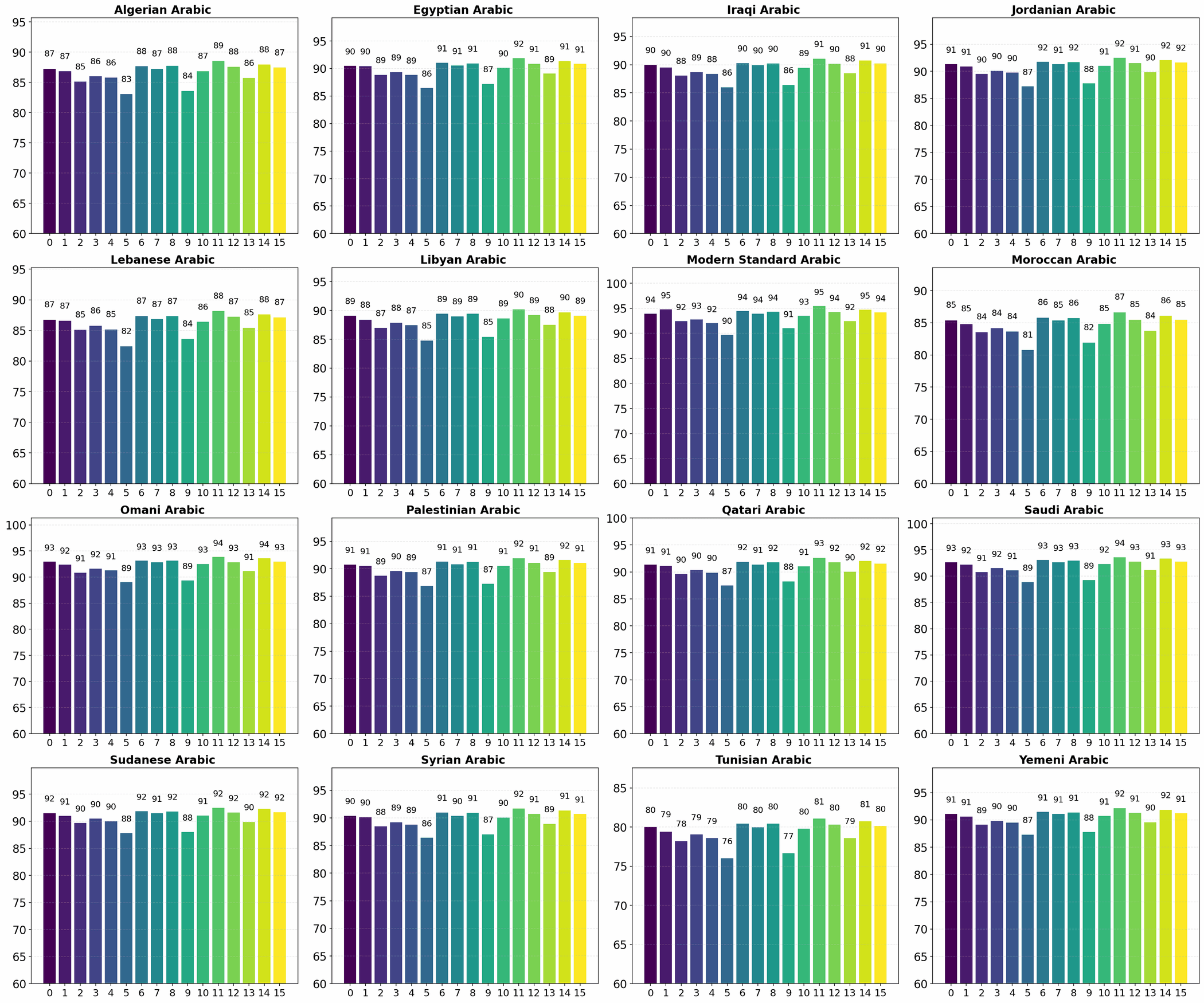}
    \caption{COMET scores \textbf{using the reference} for merged dialect test sets for all models in English to Arabic, using the \textbf{general Arabic prompt}.  See Figure \ref{fig:merged_dialects_bleu} for the mapping of the indices on the \textit{x} axis to model names. 
}
    \label{fig:comet_ref_general}
\end{figure*}

\begin{figure*}[tb]
    \centering
    \includegraphics[width=1\linewidth]{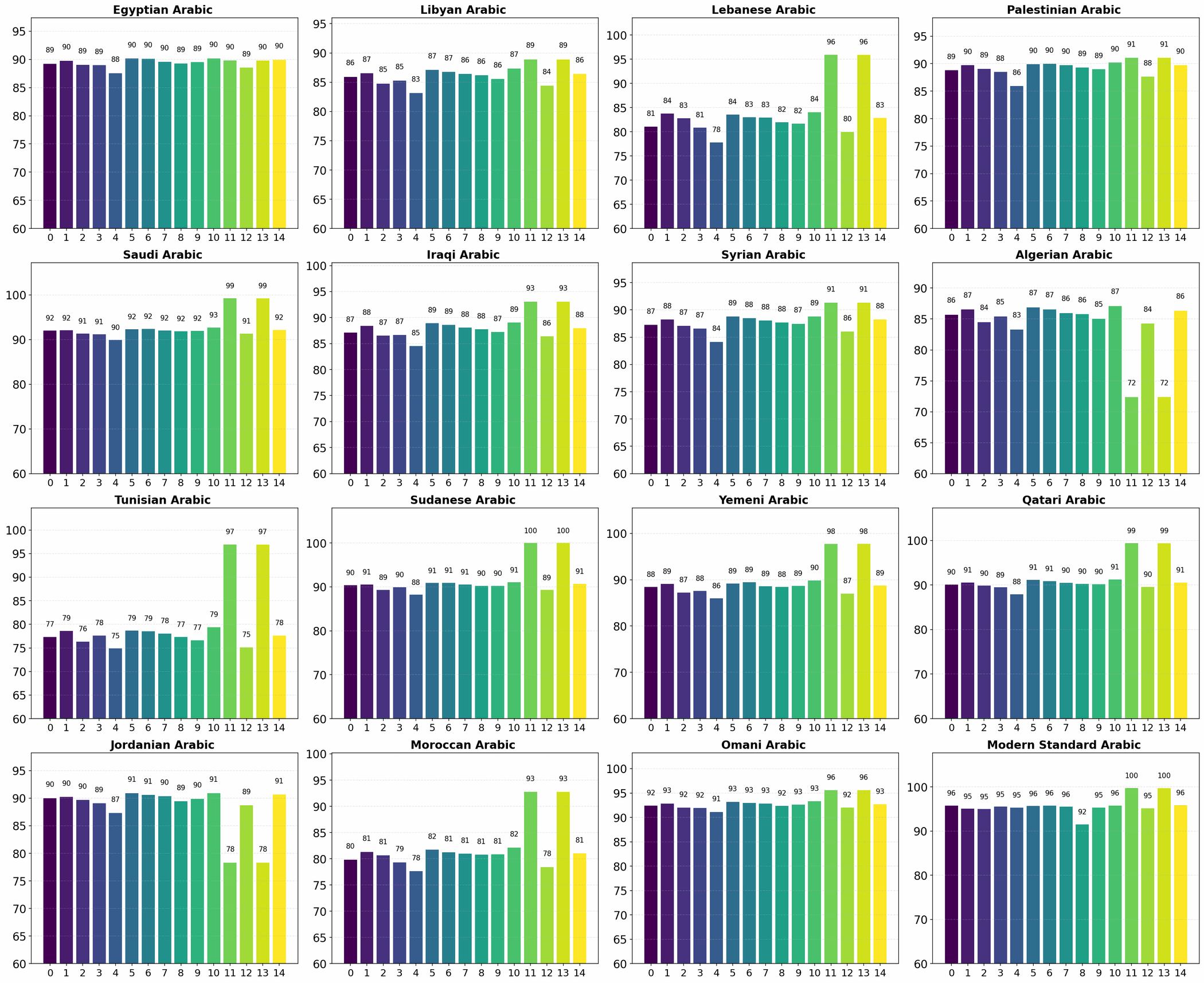}
    \caption{COMET scores without using the reference for merged dialect test sets for all models in Arabic to English.  See Figure \ref{fig:merged_dialects_bleu} for the mapping of the indices on the \textit{x} axis to model names. 
}
    \label{fig:comet_refless_aren}
\end{figure*}

\begin{figure*}[tb]
    \centering
    \includegraphics[width=1\linewidth]{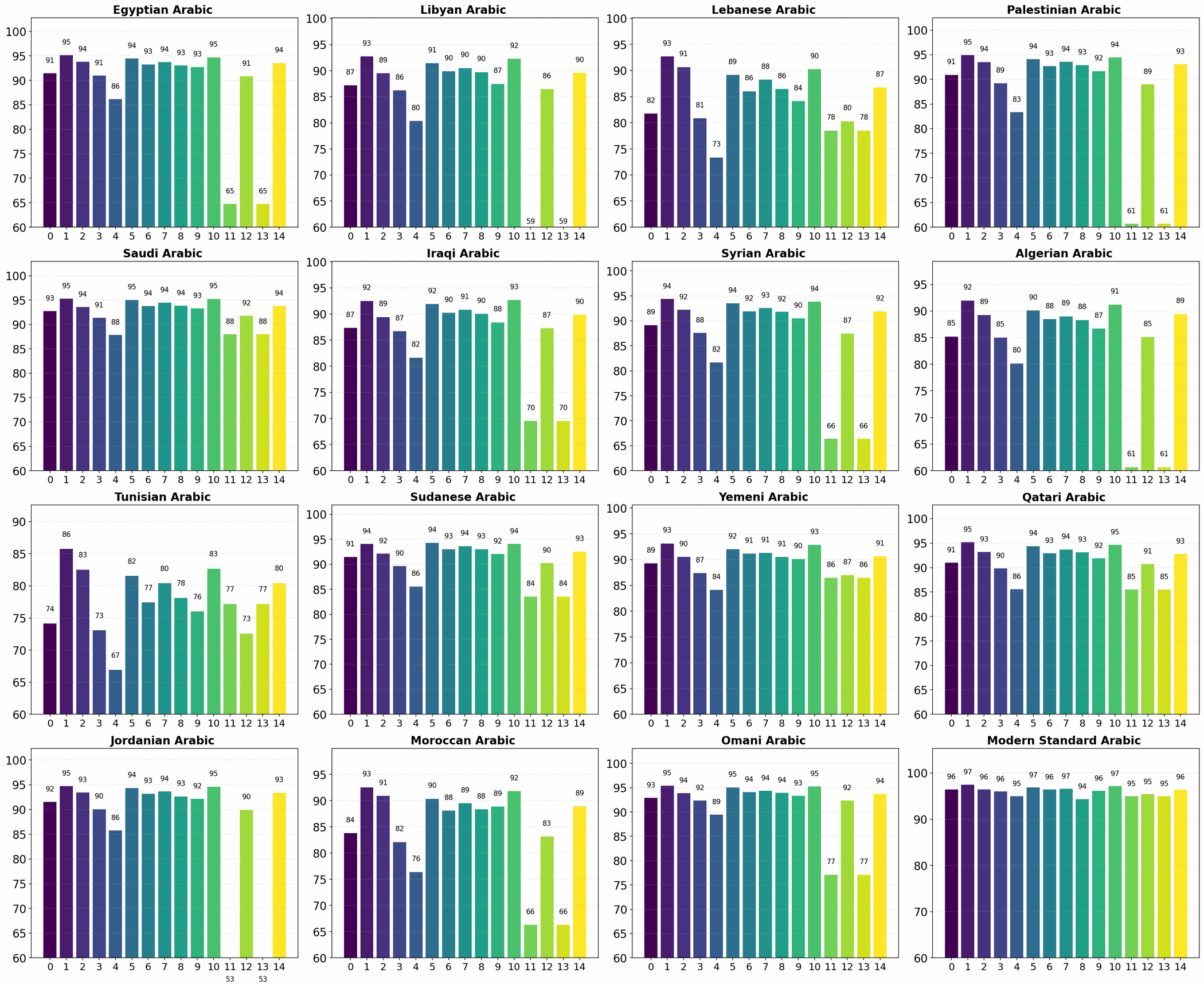}
    \caption{COMET scores \textbf{using the reference} for merged dialect test sets for all models in Arabic to English.  See Figure \ref{fig:merged_dialects_bleu} for the mapping of the indices on the \textit{x} axis to model names. 
}
    \label{fig:comet_ref_aren}
\end{figure*}

Figures \ref{fig:comet_refless_dialect}, \ref{fig:comet_ref_dialect}, \ref{fig:comet_refless_general} and \ref{fig:comet_ref_general} show reference and reference-less versions of COMET-XL on translation using the dialect-specific prompt (Figures \ref{fig:comet_refless_dialect} and \ref{fig:comet_ref_dialect}) and the general Arabic prompt  (Figures \ref{fig:comet_refless_general} and \ref{fig:comet_ref_general}). In the comparison with other metrics and our human evaluation, this shows that COMET scores actually penalize translations that are correctly realized in dialectal Arabic, due to the preference of the COMET model for MSA. Using references with COMET only partially mitigates this issue, the MSA translations are still preferred. See for example the correlations in Figure \ref{fig:correlation} for further evidence. 
For the Arabic to English direction, we show the reference-less and reference-based scores in Figures \ref{fig:comet_refless_aren} and \ref{fig:comet_ref_aren}, respectively.

\subsection{Manual Error Analysis}
The errors observed throughout the human annotation process fell into one of these four categories: 
\begin{itemize}[itemsep=0pt,topsep=0pt]
    \item \textbf{Adequacy}: This category included instances of mistranslation, 
    partial loss of source meaning, literal translation, and severely inadequate 
    or irrelevant outputs, where the meaning of the source sentence was partially 
    or completely distorted.
    
    An example of a literal translation error is the word “around” in the phrase 
    “around here,” which was translated as \RL{.hwAly} meaning “around” in the 
    sense of an approximate quantity rather than location.
    
    \item \textbf{Fluency}: Fluency errors included orthographic errors, grammatical (morphosyntactic) errors such as agreement and inflection mistakes, for example, the generated translation for How much is the breakfast? is \RL{Alf.tAr kAm} (How much breakfast) instead of \RL{Alf.tAr bkAm} with the correct proposition and meaning, other fluency errors covered awkward or non-idiomatic phrasing, where the sentence was semantically correct but sounded unnatural to a native speaker.
    \item \textbf{Dialectal appropriateness}: This category captured cases where the output did not conform to the target dialect. Errors ranged from minor cases, where only a small portion of the output was in Modern Standard Arabic (MSA), to major cases, where the output was predominantly or entirely in MSA or contained incorrect dialectal lexical choices.
    \item \textbf{Completeness}: This category included omission errors (truncated or incomplete outputs), insertion errors (addition of words not present in the source), and copying errors, such as untranslated or transliterated words, for instance the word "Airlines was teansliterated to \RL{AyrlAynz}.
\end{itemize}
Table \ref{tab:detailed_errors} shows the detailed error analysis for each model.

\begin{table*}[ht!]
\centering
\begin{tabular}{lrrrr}
\toprule
\textbf{Error Category} & \textbf{J70B} & \textbf{G-mini} & \textbf{ELLM-9B} & \textbf{CR-824} \\ 
\midrule
\textit{Adequacy} & & & & \\
Mistranslation           & 10 (66.7\%)  & 6 (21.4\%)      & 13 (13.8\%)     & 24 (34.8\%)           \\
Hallucinated translation & 1 (6.7\%)   & 0 (0.0\%)      & 0 (0.0\%)      & 1 (1.5\%)           \\
Literal Translation      & 0 (0.0\%)   & 11 (39.3\%)      & 0 (0.0\%)      & 1 (1.5\%)           \\ \addlinespace
\textit{Fluency} & & & & \\
Orthographical Error     & 1 (6.7\%)   & 1 (3.6\%)      & 6 (6.4\%)      & 0 (0.0\%)           \\
Morphosyntactic Error    & 1 (6.7\%)    & 2 (7.1\%)      & 5 (5.3\%)      & 4 (5.8\%)           \\
Non-idiomatic phrasing   & 0 (0.0\%)    & 0 (0.0\%)       & 0 (0.0\%)      & 1 (1.5\%)           \\ \addlinespace
\textit{Dialectal appropriateness} & & & & \\
Dialect Mismatch (Major) & 0 (0.0\%)    & 0 (0.0\%)      & 59 (62.8\%)      & 7 (10.1\%)           \\
Dialect Mismatch         & 0 (0.0\%)    & 7 (25.0\%)      & 10 (10.6\%)     & 30 (43.5\%)         \\ \addlinespace
\textit{Completeness} & & & & \\
Copying     & 1 (6.7\%)   & 0 (0.0\%)      & 1 (1.1\%)      & 1 (1.5\%)           \\ 
Omission    & 1 (6.7\%)   & 0 (0.0\%)      & 0 (0.0\%)       & 0 (0.0\%)           \\
Insertion     & 0 (0.0\%)    & 1 (3.6\%)      & 0 (0.0\%)       & 0 (0.0\%)           \\ 
\midrule
\textbf{Total Errors}    & 15   & 28      & 94      & 69           \\
\bottomrule
\end{tabular}
\caption{Detailed Error Analysis of Egyptian Arabic Translations generated by: Jais-2-70B-
Chat (J70B), GPT-4.1-mini (G-mini), EuroLLM-9B-
Instruct (ELLM-9B), Command-r-08-2024 (CR-824).}
\label{tab:detailed_errors}
\end{table*}
\subsection{Dialect classification}

\begin{figure*}[tb]
    \centering
    \includegraphics[width=1\linewidth]{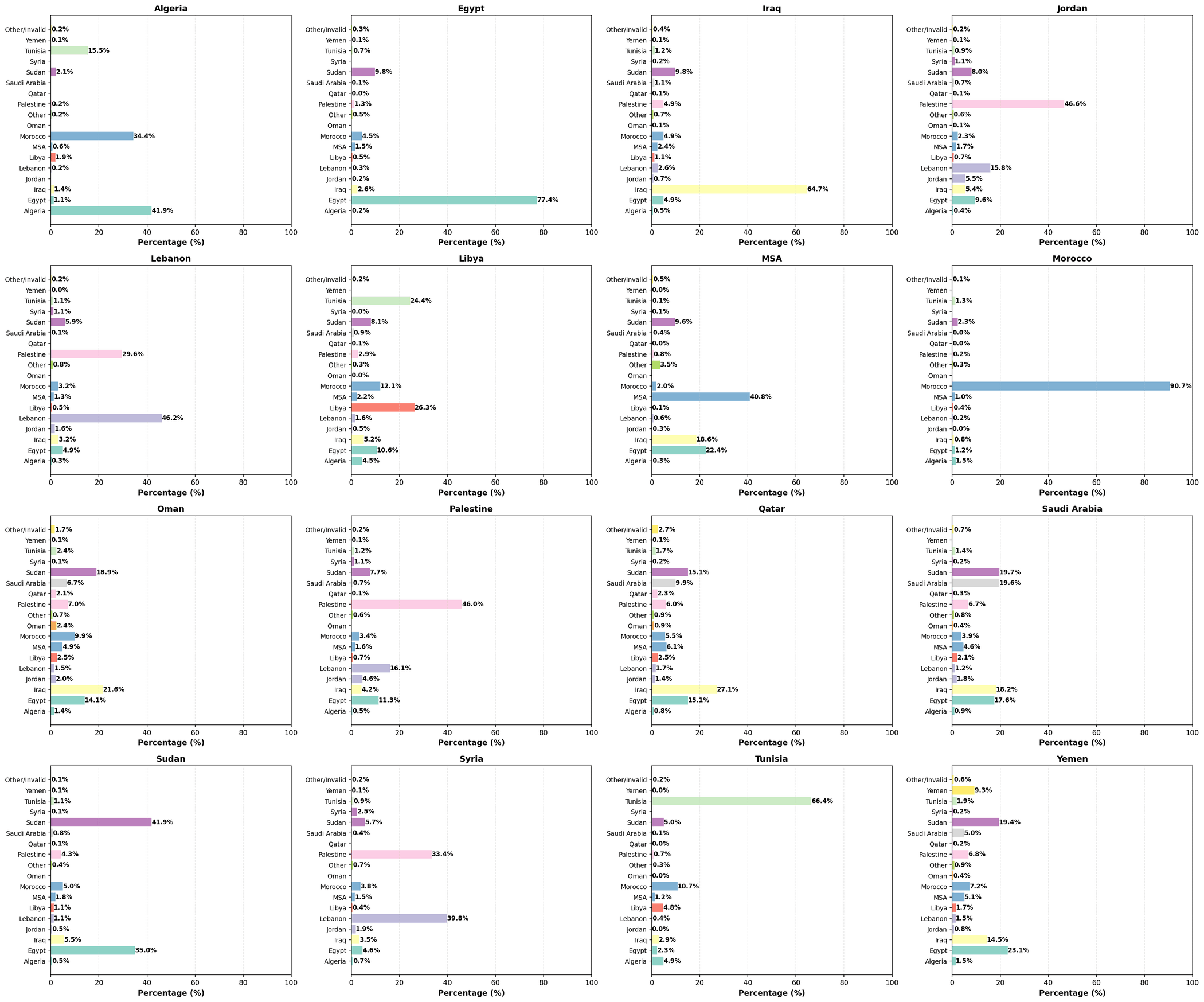}
     \caption{Distribution of dialects in translations produced by \texttt{Jais-2-70B-Chat} using the dialect-specific prompt. The dialects are classified on the sentence level, using \texttt{Jais-2-70B-Chat}.}
    \label{fig:dialect_jais_by_jais}
\end{figure*}

\begin{figure*}[tb]
    \centering
    \includegraphics[width=1\linewidth]{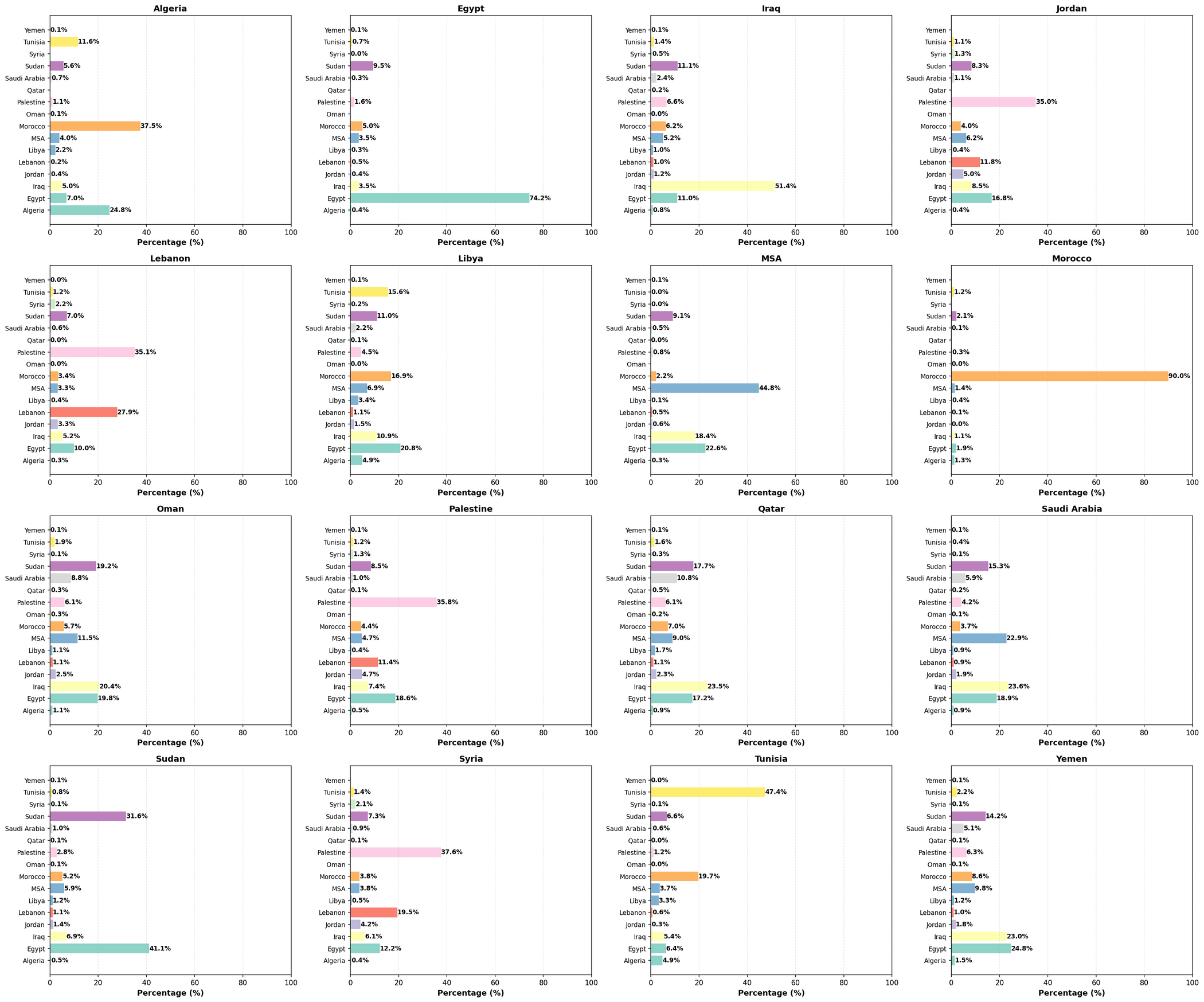}
    \caption{Distribution of dialects in translations produced by \texttt{gpt-4.1-mini} using the dialect-specific prompt. The dialects are classified on the sentence level, using \texttt{Jais-2-70B-Chat}.}
    \label{fig:dialect_gpt_mini_by_jais}
\end{figure*}

\begin{figure*}[tb]
    \centering
    \includegraphics[width=1\linewidth]{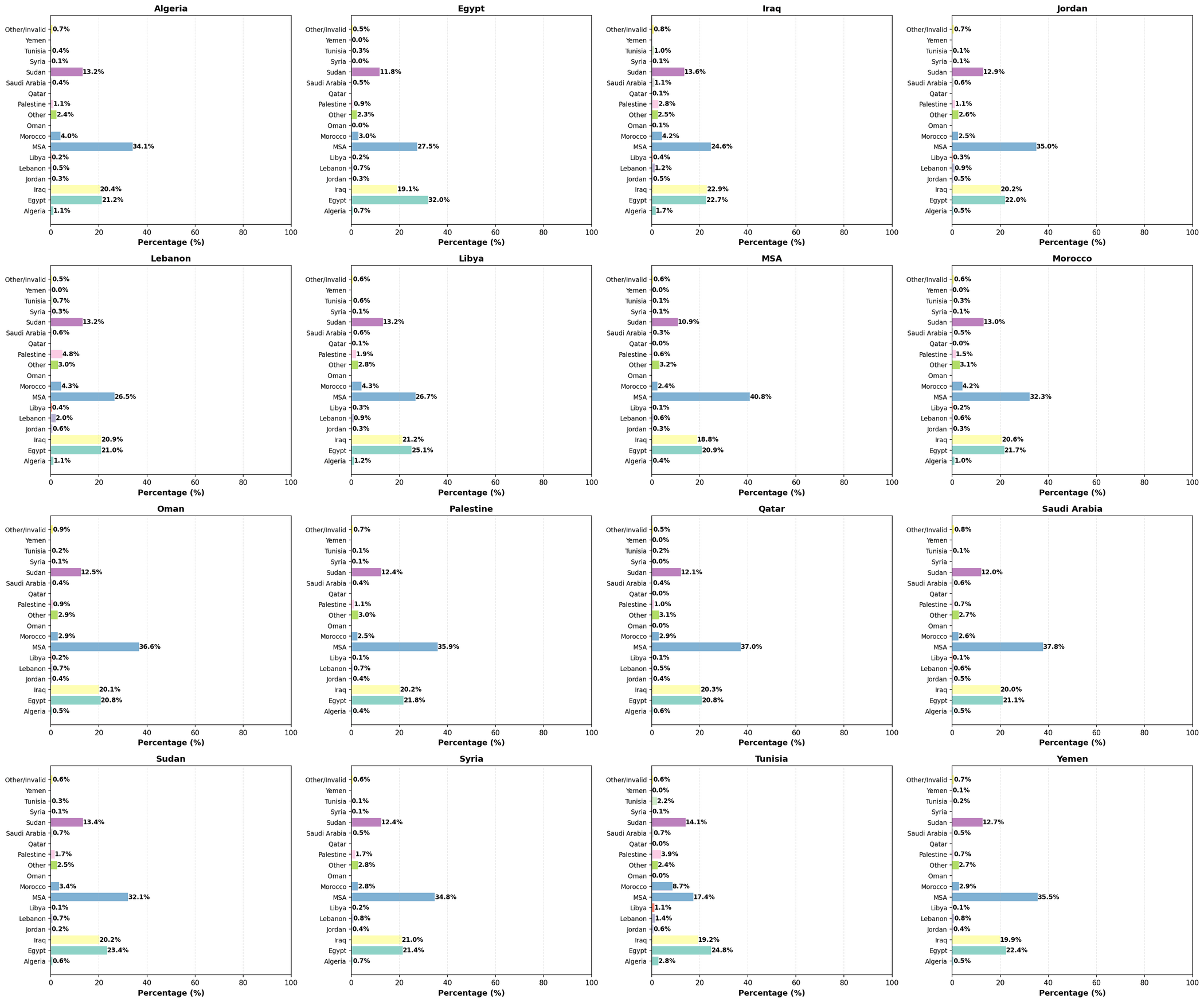}
     \caption{Distribution of dialects in translations produced by \texttt{EuroLLM-9B-Instruct} using the dialect-specific prompt. The dialects are classified on the sentence level, using \texttt{Jais-2-70B-Chat}.}
    \label{fig:dialect_ellm_by_jais}
\end{figure*}

\begin{figure*}[tb]
    \centering
    \includegraphics[width=1\linewidth]{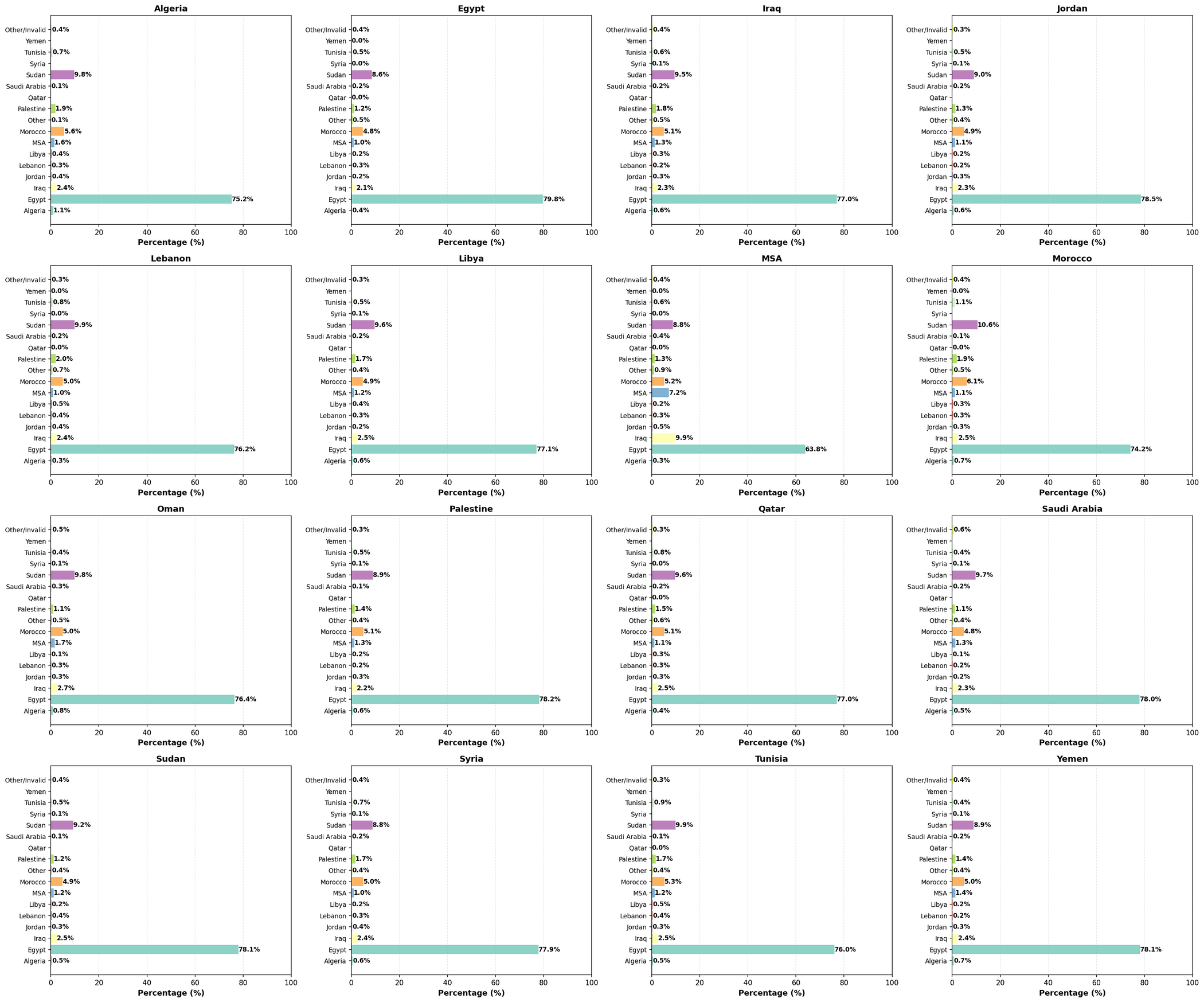}
    \caption{Distribution of dialects in translations produced by \texttt{Nile-Chat-12B} using the dialect-specific prompt. The dialects are classified on the sentence level, using \texttt{Jais-2-70B-Chat}.}
    \label{fig:dialect_nile_by_jais}
\end{figure*}

\begin{figure*}[tb]
    \centering
    \includegraphics[width=1\linewidth]{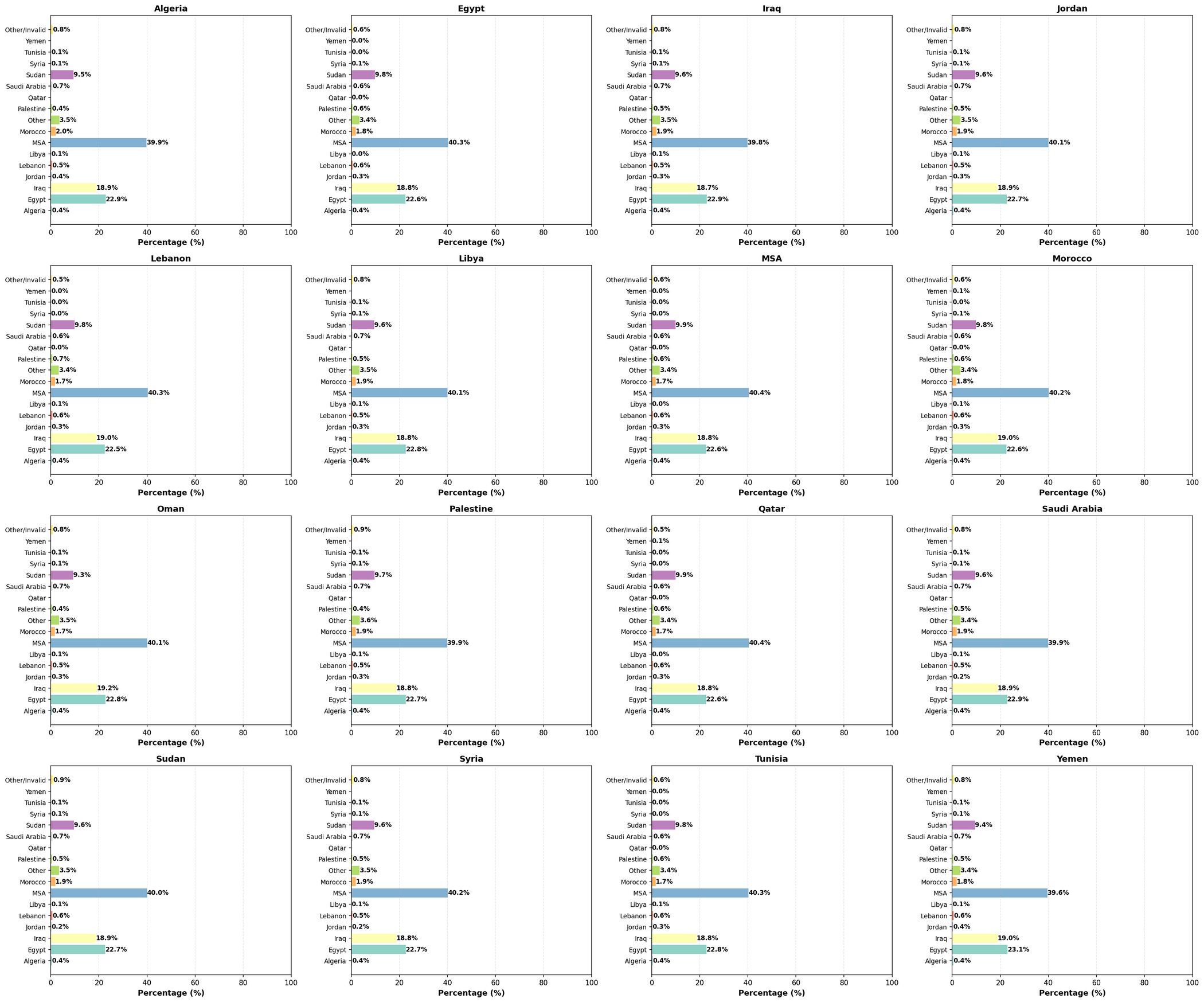}
    \caption{Distribution of dialects in translations produced by \texttt{Jais-2-70B-Chat} using the general Arabic prompt. The dialects are classified on the sentence level, using \texttt{Jais-2-70B-Chat}.}
    \label{fig:dialect_general_jais_by_jais}
\end{figure*}

\begin{figure*}[tb]
    \centering
    \includegraphics[width=1\linewidth]{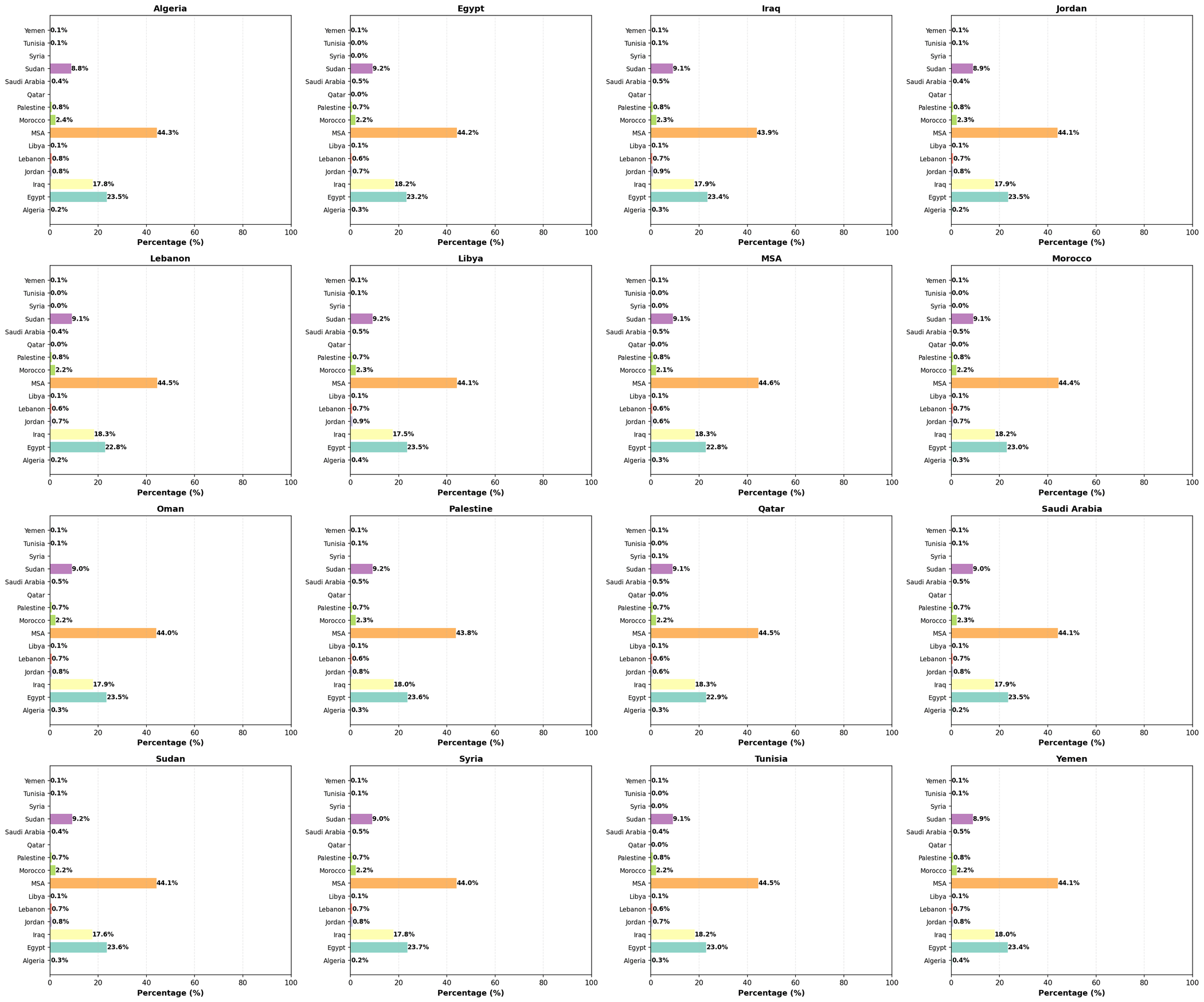}
     \caption{Distribution of dialects in translations produced by \texttt{gpt-4.1-mini} using the general Arabic prompt. The dialects are classified on the sentence level, using \texttt{Jais-2-70B-Chat}.}
    \label{fig:dialect_general_gpt_mini_by_jais}
\end{figure*}

\begin{figure*}[tb]
    \centering
    \includegraphics[width=1\linewidth]{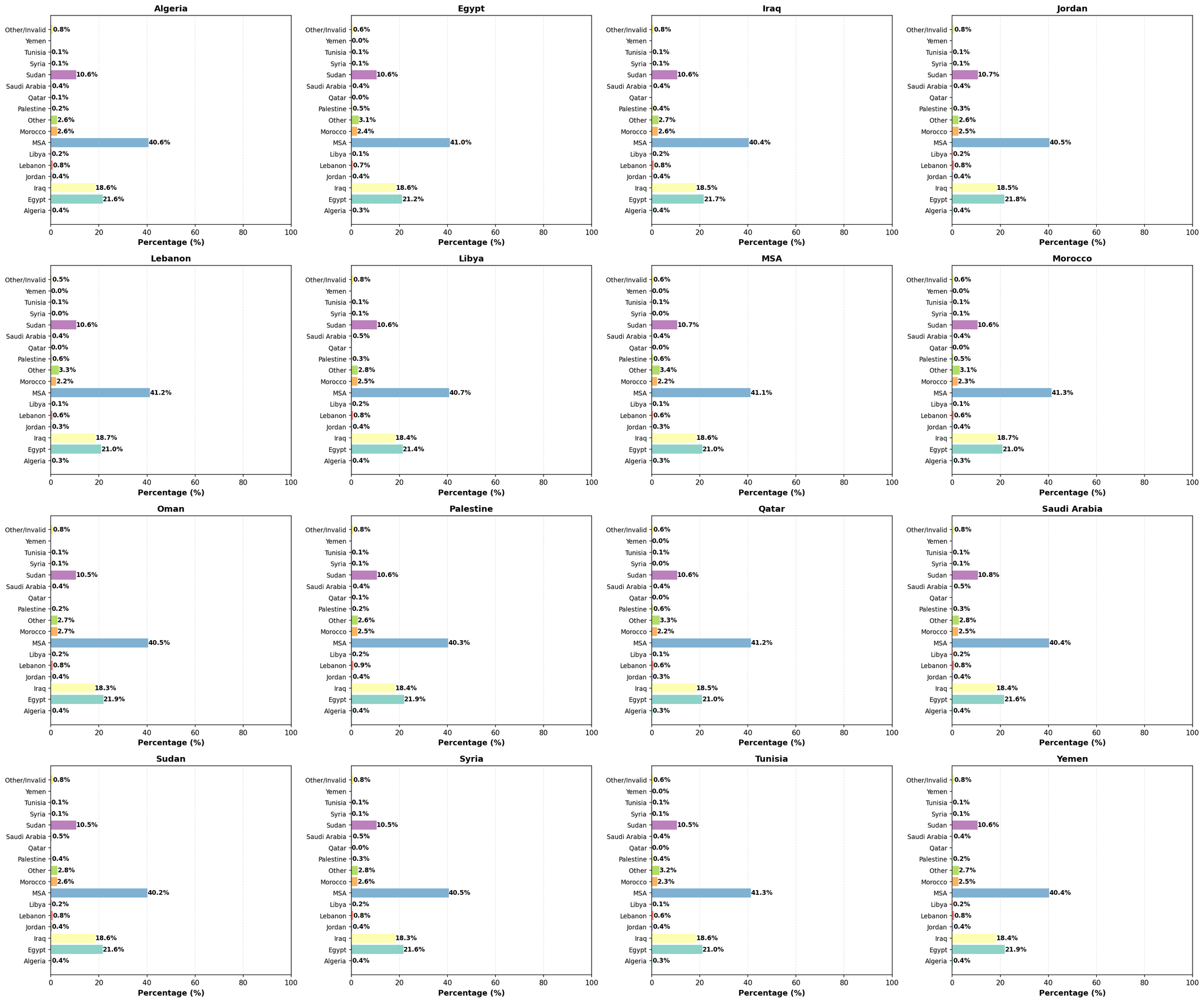}
    \caption{Distribution of dialects in translations produced by \texttt{EuroLLM-9B-Instruct} using the general Arabic prompt. The dialects are classified on the sentence level, using \texttt{Jais-2-70B-Chat}.}
    \label{fig:dialect_general_ellm_by_jais}
\end{figure*}

\begin{figure*}[tb]
    \centering
    \includegraphics[width=1\linewidth]{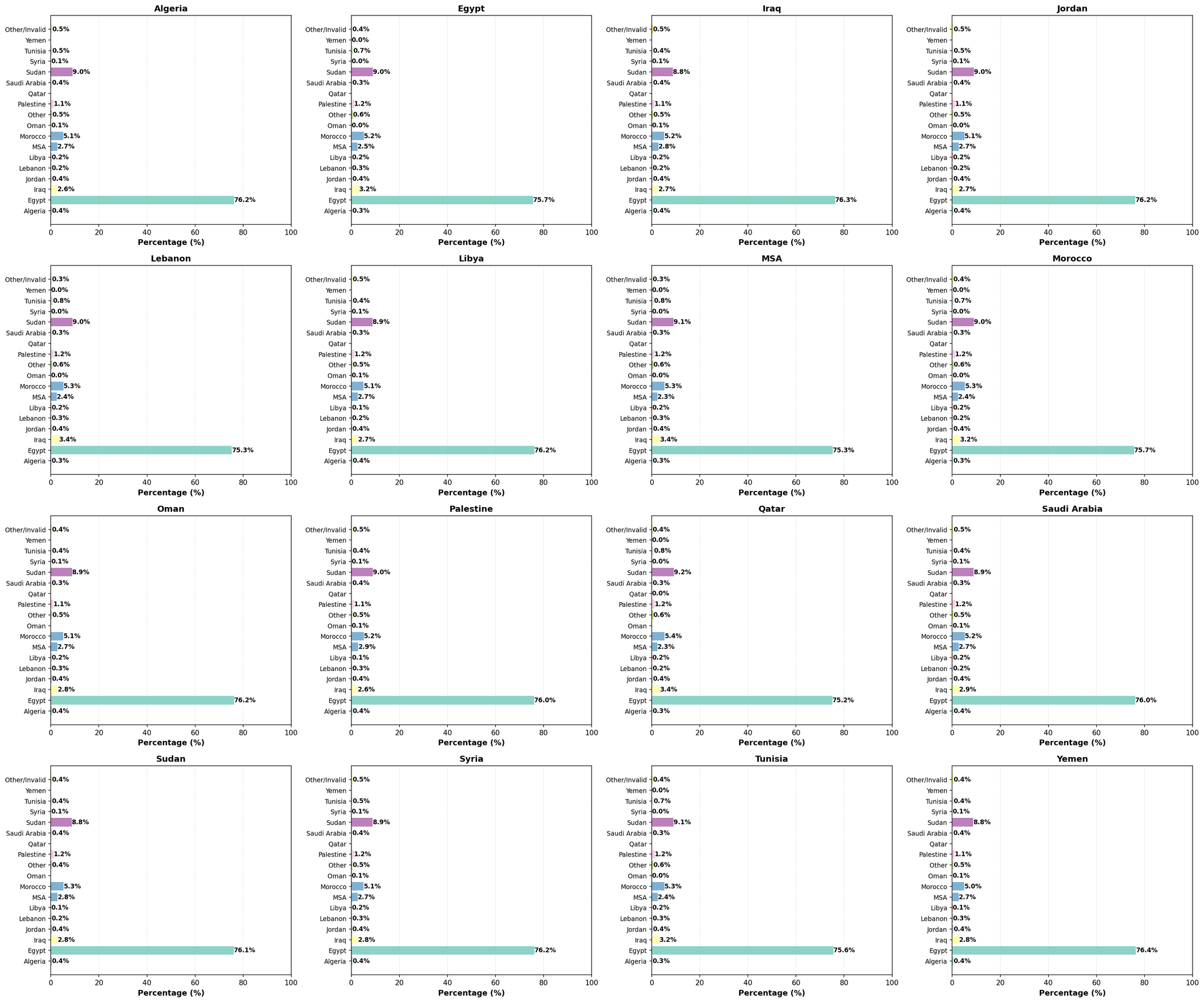}
    \caption{Distribution of dialects in translations produced by \texttt{Nile-Chat-12B} using the general Arabic prompt. The dialects are classified on the sentence level, using \texttt{Jais-2-70B-Chat}.}
    \label{fig:dialect_general_nile_by_jais}
\end{figure*}

\begin{figure*}[tb]
    \centering
    \includegraphics[width=1\linewidth]{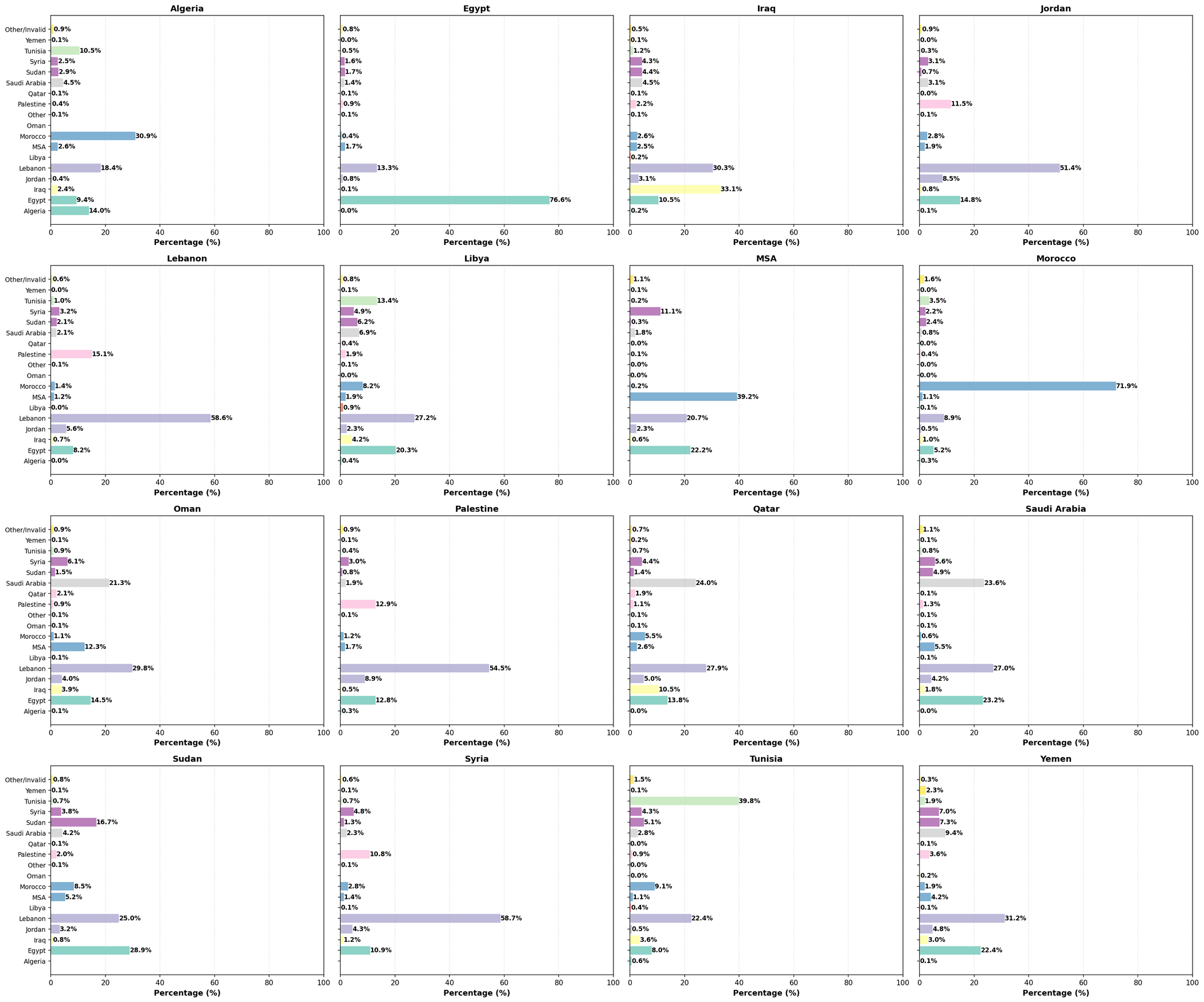}
     \caption{Distribution of dialects in the reference translations. The dialects are classified on the sentence level, using \texttt{gemma-3-27b-it}.}
    \label{fig:dialect_reference_by_gemma}
\end{figure*}

\begin{figure*}[tb]
    \centering
    \includegraphics[width=1\linewidth]{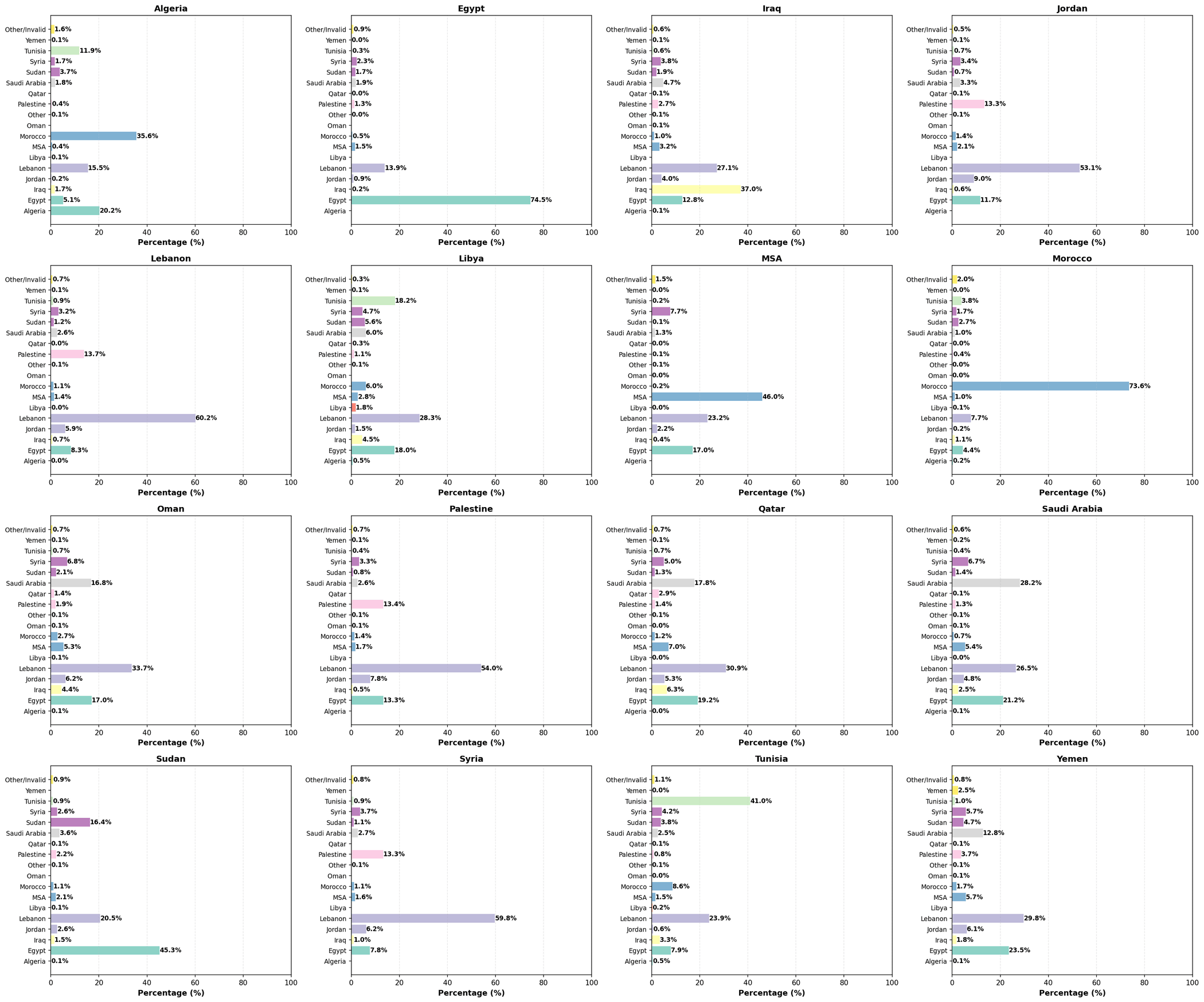}
     \caption{Distribution of dialects in translations produced by \texttt{Jais-2-70B-Chat} using the dialect-specific prompt. The dialects are classified on the sentence level, using \texttt{gemma-3-27b-it}.}
    \label{fig:dialect_jais_by_gemma}
\end{figure*}

\begin{figure*}[tb]
    \centering
    \includegraphics[width=1\linewidth]{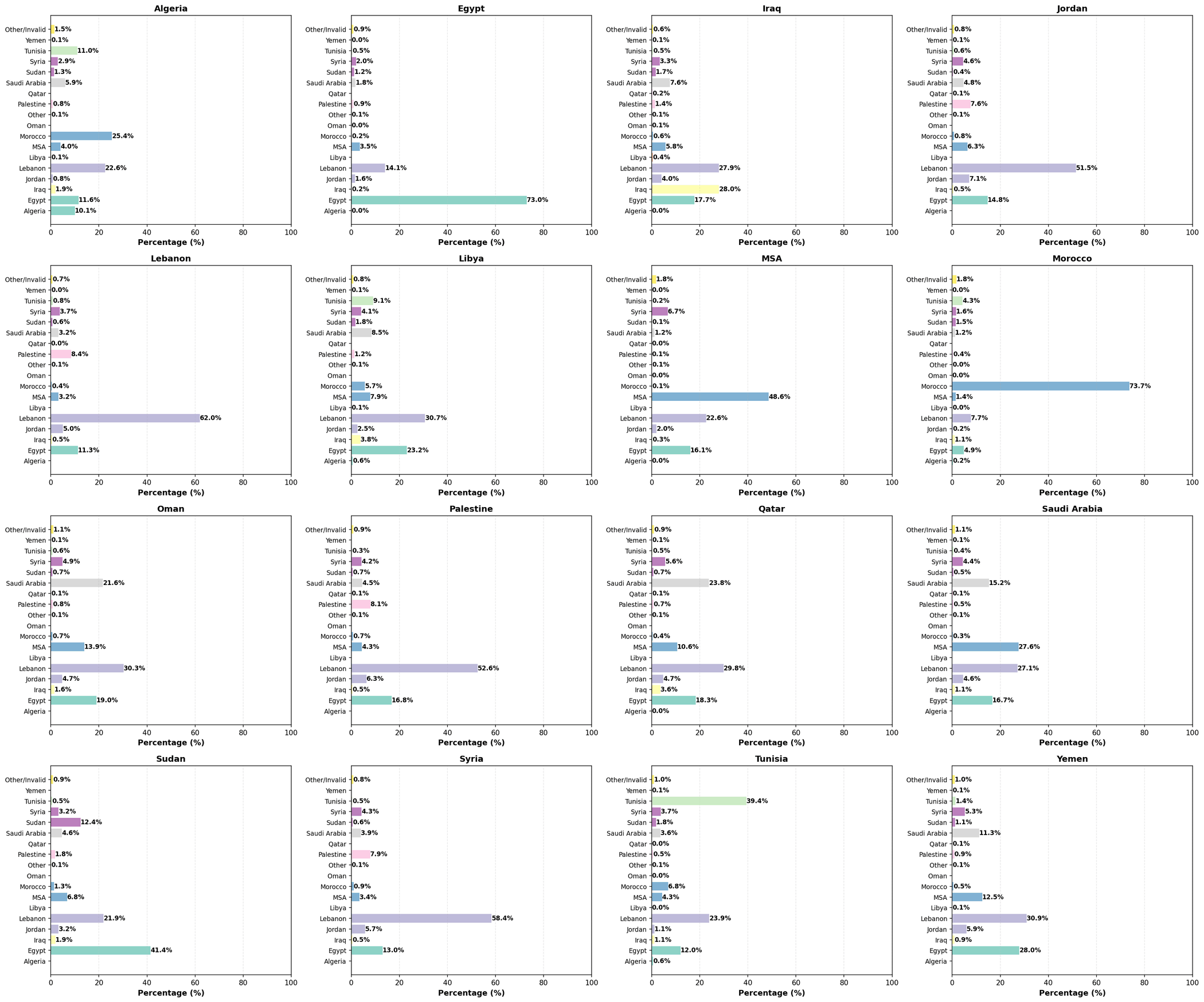}
    \caption{Distribution of dialects in translations produced by \texttt{gpt-4.1-mini} using the dialect-specific prompt. The dialects are classified on the sentence level, using \texttt{gemma-3-27b-it}.}
    \label{fig:dialect_gpt_mini_by_gemma}
\end{figure*}

\begin{figure*}[tb]
    \centering
    \includegraphics[width=1\linewidth]{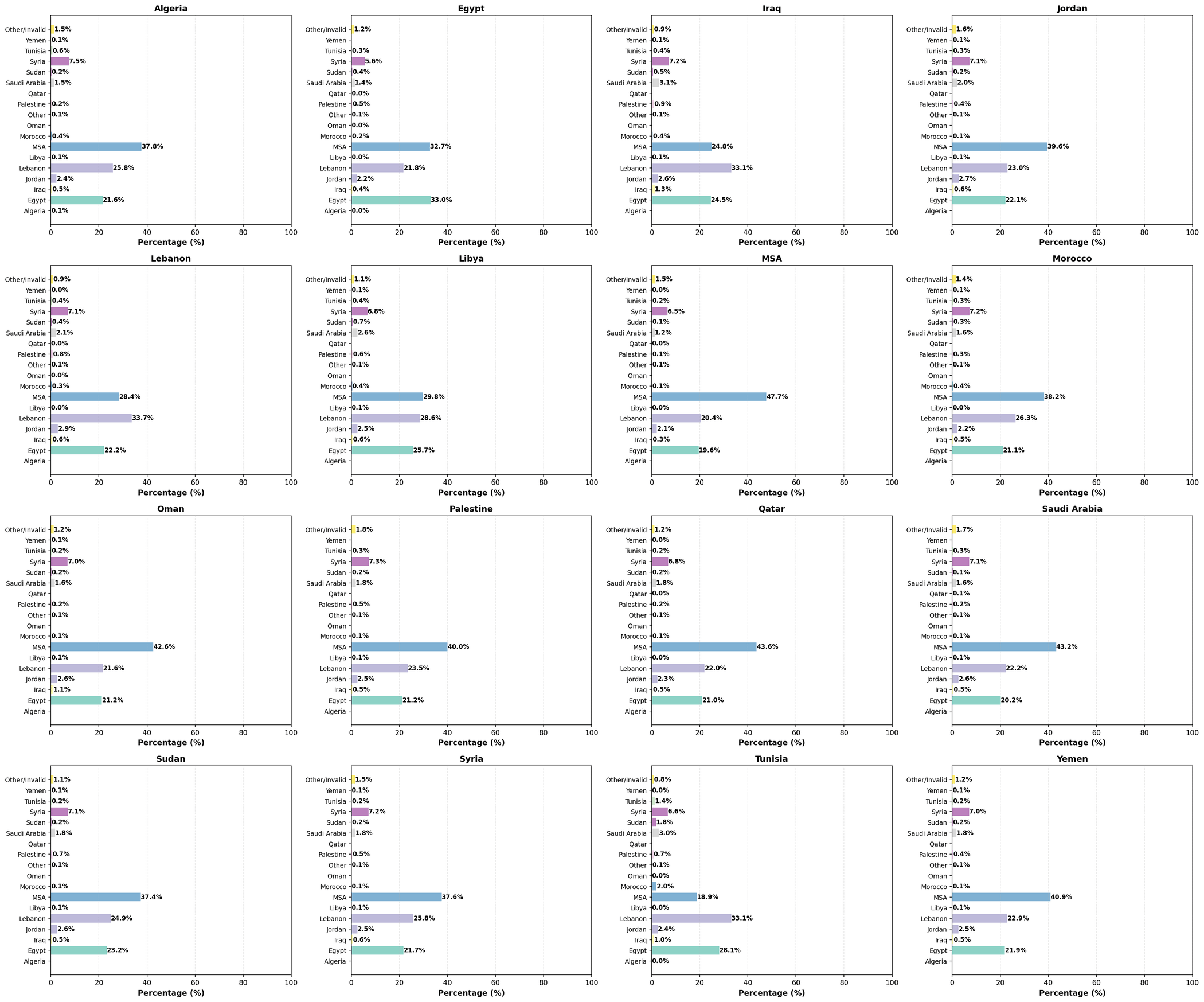}
     \caption{Distribution of dialects in translations produced by \texttt{EuroLLM-9B-Instruct} using the dialect-specific prompt. The dialects are classified on the sentence level, using \texttt{gemma-3-27b-it}.}
    \label{fig:dialect_ellm_by_gemma}
\end{figure*}

\begin{figure*}[tb]
    \centering
    \includegraphics[width=1\linewidth]{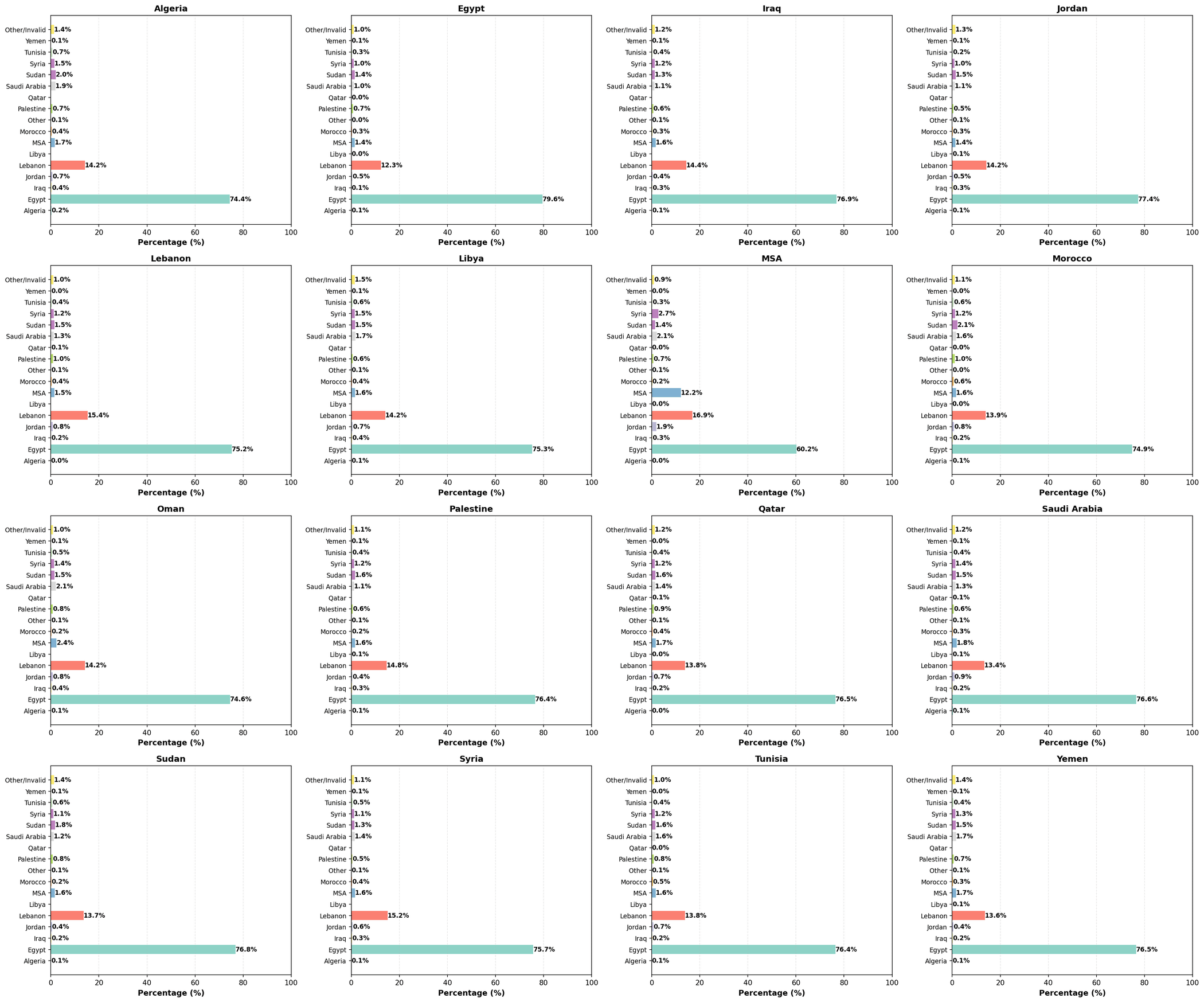}
    \caption{Distribution of dialects in translations produced by \texttt{Nile-Chat-12B} using the dialect-specific prompt. The dialects are classified on the sentence level, using \texttt{gemma-3-27b-it}.}
    \label{fig:dialect_nile_by_gemma}
\end{figure*}

Figures \ref{fig:dialect_jais_by_jais}, \ref{fig:dialect_gpt_mini_by_jais}, \ref{fig:dialect_ellm_by_jais} and \ref{fig:dialect_nile_by_jais} show dialect classification performed by the \texttt{Jais-2-70B-Chat} for test sets translated using the dialect specific prompt by \texttt{Jais-2-70B-Chat}, \texttt{gpt-4.1-mini}, \texttt{EuroLLM-9B-Instruct} and \texttt{Nile-Chat-12B}, respectively. We can see a number of interesting trends. First of all, we see that according to the assessment by the classification model, Jais-2-70B-Chat and gpt-4.1-mini produce outputs that have similar distribution dialects as the reference\XXX{TODO, check if true}. It is apparent that they reflect the dialect name in the prompt, for comparison, see Figures \ref{fig:dialect_general_jais_by_jais}, \ref{fig:dialect_general_gpt_mini_by_jais}, \ref{fig:dialect_general_ellm_by_jais}, \ref{fig:dialect_general_nile_by_jais}, which present the classification results on translation produced by the same models general Arabic prompt instead of dialect-specific one. The appropriate dialects are notably more represented in the translations with the dialect-specific prompt. 
On the other hand, the EuroLLM model is not producing the texts in the correct dialects, using mostly MSA instead. This tendency was also shown by our manual evaluation. Another observation is related to the Nile-Chat-12B model, which, regardless of dialect specification, produces texts in the Egyptian and Sudanese dialects. This is intended by the model authors and our evaluation thus confirms that the model is Nile-dialects specific.

Figures \ref{fig:dialect_jais_by_gemma}, \ref{fig:dialect_gpt_mini_by_gemma}, \ref{fig:dialect_ellm_by_gemma} and \ref{fig:dialect_nile_by_gemma} present the distributions of dialects in the translations performed by the same set of models as above, using the dialect-specific prompt, but classified using \texttt{gemma-27b-it} instead of \texttt{Jais-2-70B-Chat}. Figure \ref{fig:dialect_reference_by_gemma} shows the dialect distribution in the reference translations, also assessed by the \texttt{gemma-27b-it} model. It is apparent that different LLMs have different priors about the dialect distributions, e.g. \texttt{gemma-27b-it} classifies noticeably larger portion of the texts as Lebanese, even for test sets from other dialects. 

\label{sec:app_dial}

\section{Prompts}
\label{sec:prompts}
\subsection{Translation Prompt}
\textbf{System prompt}
\begin{tcolorbox}[promptbox]
You are a professional, very precise translator and a native speaker.
Translate inputs based on the instructions and always print out only
the text of the best possible translation, with no explanations.

Keep the same formatting (e.g.\ markup, lines, spacing) as the original.
Do not translate untranslatable parts of the input
(URLs, code, and similar).
\end{tcolorbox}
\textbf{Instruction prompt}
\begin{tcolorbox}[promptbox]
Translate the following text into \{lang\}, only print out the translation, not add any explanations: \{line\}
\end{tcolorbox}
\subsection{GEMBA Translation Evaluation Prompt}

\begin{tcolorbox}[promptbox]
Score the following translation from \{source\_lang\} to \{target\_lang\}
on a continuous scale from 0 to 100, where a score of zero means
``no meaning preserved'' and a score of one hundred means
``perfect meaning and grammar''.

Be very strict with checking the target language or dialect.
For example, if an Arabic dialect is specified and the translation
is in Modern Standard Arabic or another dialect, penalize harshly
by subtracting 50 points from the score you would otherwise give,
if the dialect is completely incorrect.

\{source\_lang\} source: ``\{source\_seg\}''

\{target\_lang\} translation: ``\{target\_seg\}''

Score:
\end{tcolorbox}

\subsection{Dialect Classification Prompt}

\begin{tcolorbox}[promptbox]
You are an expert in Arabic dialects. Classify the following Arabic text
by determining which country or region it is from.

Choose the most appropriate country from this list:
\{countries\_str\},
or ``Modern Standard Arabic'' if it is formal standard Arabic,
or ``Other'' if none of the above apply.

Arabic text: ``\{text\}''

Respond with ONLY the country name from the list above, nothing else.
\end{tcolorbox}

\end{document}